\pdfoutput=1
\documentclass[11pt]{article}

\usepackage[]{emnlp2021}

\usepackage{times}
\usepackage{latexsym}

\usepackage[T1]{fontenc}
\usepackage[utf8]{inputenc}

\usepackage{microtype}
\usepackage{graphicx}
\usepackage{makecell}
\usepackage{caption}
\usepackage{subcaption}
\usepackage{tabularx}
\usepackage{ltablex}
\usepackage{makecell}
\usepackage{multirow}
\usepackage{relsize}

\usepackage{float}
\restylefloat{table}

\title{Efficient Multi-Modal Embeddings from Structured Data}

\author{Anita L. Verő \\
  Department of Computer Science\\
  and Technology\\
  University of Cambridge\\
  \texttt{alv34@cam.ac.uk}\\
  \And
  Ann Copestake \\
  Department of Computer Science\\
  and Technology\\
  University of Cambridge\\
  \texttt{aac10@cl.cam.ac.uk}}

\begin{document}
\maketitle
\begin{abstract}
Multi-modal word semantics aims to enhance embeddings with perceptual input, assuming that human meaning representation is grounded in sensory experience.
Most research focuses on evaluation involving direct visual input, however, visual grounding can contribute to linguistic applications as well. Another motivation for this paper is the growing need for more interpretable models and for evaluating model efficiency regarding size and performance.
This work explores the impact of visual information for semantics when the evaluation involves no direct visual input, specifically semantic similarity and relatedness.
We investigate a new embedding type in-between linguistic and visual modalities, based on the structured annotations of Visual Genome.
We compare uni- and multi-modal models including structured, linguistic and image based representations.
We measure the efficiency of each model with regard to data and model size, modality / data distribution and information gain. The analysis includes an interpretation of embedding structures.
We found that this new embedding conveys complementary information for text based embeddings. It achieves comparable performance in an economic way, using orders of magnitude less resources than visual models.
\end{abstract}

\section{Introduction}

\begin{figure}[ht]
\centering
\includegraphics[width=\linewidth]{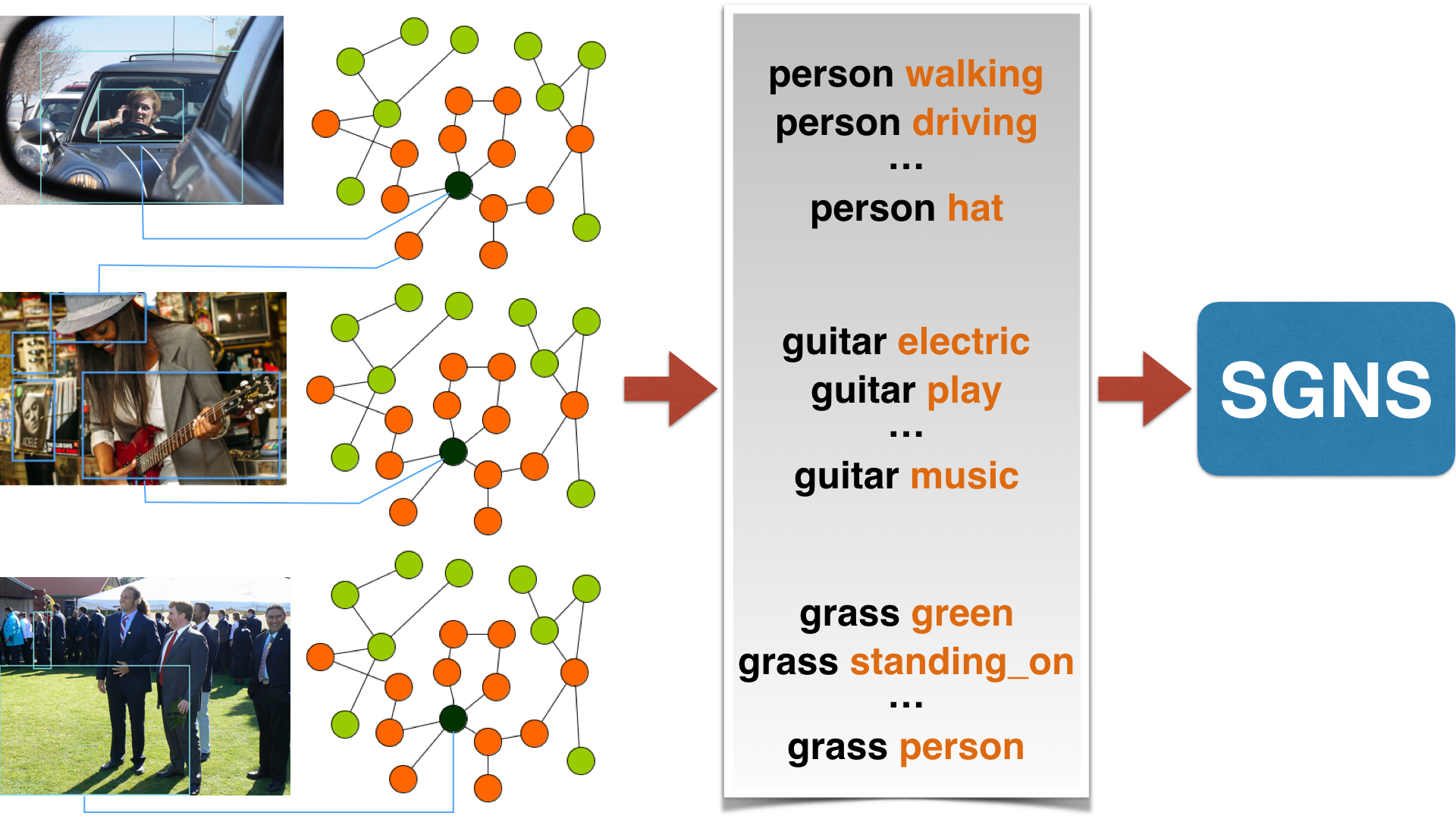}
\caption{Generating contexts for embeddings from Visual Genome Scene-graphs. The context words (orange) used are up to three links from a target node (black). Photos are from \url{https://visualgenome.org/}}
\label{f:vg_context}
\end{figure}

Multi-modal distributional semantics is inspired by the hypothesis that human semantic representation depends on sensorimotor experience (Symbol grounding problem \citep{harnad1990symbol}), and the idea in cognitive science that meaning in the brain is represented both visually and verbally (Dual coding theory \citep{bucci1985dual}).

Most work focuses on evaluation tasks (and therefore on models which perform well on them) involving direct visual input, such as Visual Question Answering \citep{antol2015vqa, srivastava2012multimodal,kiros2014multimodal,socher2014grounded,tsai2019multimodal,lu2019vilbert,su2019vl,majumdar2020improving}. In these cases the usefulness of visual input is not surprising. However, visual common sense can be useful in purely linguistic tasks as well \citep{bruni2014multimodal, Kiela:2014emnlp,lazaridou2015combining}. The analysis of the reasons has been addressed by fewer work \citep{kiela2014improving, Kiela:2016emnlp, davis2019deconstructing}, however, more research is needed on  the conditions and the type of information visual modalities contribute.
Another motivation for this paper is the growing need for more interpretable models and to study model efficiency regarding size and performance \citep{bender2021dangers}.

This work explores the impact of visual information in meaning representations for semantic similarity and relatedness (Sec.~\ref{sec:eval_exp}).
We investigate a new, visually structured, textual embedding (Sec.~\ref{sec:emb_types}) in-between linguistic and visual modalities, based on the Visual Genome (VG) Scene-graphs as a structured data-source \citep{krishnavisualgenome}.

Our main contribution is twofold.
First, we measure the efficiency of each model with regard to data and model size and modality / data distribution (Sec.~\ref{exp:size_perf}). 
Secondly, we analyse the interpretation of multi-modal models in terms of the Information Gain of modalities and embedding cluster structures (Sec.~\ref{exp:interpret}).
Sec.~\ref{sec:eval_exp} includes the description of evaluation. Experiments and discussion of the results are presented in Sec.~\ref{exp:size_perf} and \ref{exp:interpret}. The code for all the experiments is openly available\footnote{\url{https://github.com/anitavero/embeval}}.

We found that our structured model conveys complementary information for text based embeddings. It achieves comparable or better performance in an economic way, by using orders of magnitude less resources than visual models.
An important future application would be using automatically generated scenes graphs \citep{xu2020survey}, mitigating the limitation of this approach, which is the manual labour required for creating VG. This would serve as a highly effective tool for low resource languages.

\section{Related Work}

Our VG Scene-graph based model (Sec.~\ref{sec:emb_types}) is closely related to \citealp{kuzmenko2019distributional,herbelot2020re}. 
Even though our model is simpler (does not involve a truth theoretic model), our experiments support their claim about the efficiency of this data-source, while also showing its economic advantages in multi-modal models.

Our control method for data size and frequency ranges (Sec.~\ref{exp:size_perf}) is inspired by the text-based case of \citealp{sahlgren2016effects}. Our results support theirs, while going far beyond in the amount of text data, besides adding other modalities.

\section{Embedding Types} \label{sec:emb_types}

We define three different types of embeddings, which vary in training data type and learning algorithm.
\textit{Linguistic Embeddings} ($E_L$) are vector spaces which are learnt by an algorithm trained on large text data. 
% It can be any of the standard neural models. Here we use Skip-Gram with Negative Sampling (SGNS) \citep{mikolov2013efficient, mikolov2013distributed}.
% In this work we use three FastText \citealp{mikolov2018advances}: \textit{wiki-news-300d-1M}, \textit{wiki-news-300d-1M-subword}, \textit{crawl-300d-2M} and a Skip-Gram model (SGNS) \citealp{mikolov2013efficient, mikolov2013distributed}. For their description see App~\ref{sec:models}.
\textit{Visual Embeddings} ($E_V$) consist of vectors which have been trained on images associated to words. The learning algorithm is typically a Convolutional Neural Network \citep{lecun1989backpropagation}.
In this paper we present results using Skip-Gram with Negative Sampling (SGNS) \citep{mikolov2013efficient, mikolov2013distributed} for $E_L$, due to its simplicity and efficiency, trained on a 2020 English Wikipedia dump. We ran a feedforward step of \textit{ResNet-152} \citep{he2016deep} on Google Images for $E_V$\footnote{We apply mean aggregation on the first 10 image results which has been found the best performing in \citealp{Kiela:2016emnlp}.}, which came out best performing in our preliminary experiments, where we compared four linguistic, five visual and our structured models. (Details are in App.~\ref{app:compare}.)
 % (here we use \textit{ResNet-152} \citep{he2016deep}), which has a specified architecture for learning abstract patterns from image data. 

% \paragraph{CNN Models}
% We use \textit{AlexNet} \citealp{Krizhevsky:2012nips} (won the ILSVRC 2012 challenge), which performed best in \citealp{Kiela:2016emnlp} and compare it to \textit{ResNet-152} architecture \citealp{he2016deep}, pretrained for the ILSVRC 2015 challenge. 

% \paragraph{Image Sources}
% Following \citealp{Kiela:2016emnlp} we run a CNN feed-forward step on \textit{Google Images} for each test word.
% We compare this big data source to visual representations obtained for the densely annotated \textit{Visual Genome (VG) Images} \citealp{krishnavisualgenome} (see App~\ref{sec:VG} Fig~\ref{f:scene_graphs}), either on the \textit{internal} object bounding box images or on the \textit{whole} images, similarly to \citealp{davis2019deconstructing}.
% After mapping images to a vector space, we need a method which associates one vector to a word. 
% We usually have multiple image results for a word, hence this method has to be a vector aggregation. We apply mean aggregation which has been found the best performing one \citealp{Kiela:2016emnlp}. They found that after 10-20 images the performance plateaus across the board, therefore, in this study we always use 10 images for each word representation.

\textit{Structured Embeddings} ($E_S$) are in-between visual and linguistic, trained on visually structured text.
We exploit the dense VG image annotations represented as scene-graphs of objects, attributes, and pairwise relationships\footnote{Illustration: \url{https://visualgenome.org/}}. We introduce a new method for generating contexts from VG Scene-graphs, which is the input for an SGNS. Context words are extracted by taking $r$ steps around a target node (word) in a breath first search manner (Fig.~\ref{f:vg_context}), where radius $r$ is a parameter. In this work we present results using $r=3$, which we found best performing.
% The collection of these datasets usually involve some manual design and labour, therefore they are much smaller in terms of the used computer memory in bytes. VG images are annotated with region graph representations of objects, attributes, and pairwise relationships. Each of these region graphs are combined to form a scene graph with all the objects grounded to the image. We use the scene-graph annotations as a corpus and train a SGNS algorithm, where the context corresponds to a radius in this graph around an object or predicate node (\textit{VG SceneGraph}). The radius is the number of steps we take starting from a node in a breadth first search manner\footnote{In this work we present results using a radius of three, which we found best performing}. The context words are all the node labels within this sub-graph, see Fig.~\ref{f:vg_context}\footnote{Photos are borrowed from \url{http://visualgenome.org/}}.
% This model is linguistic in a sense that it only uses text context in the graph neighbourhood, without grounding it to visual features. However, it still uses visual information implicitly, since the graph represents relationships in visual scenes.

\section{Evaluation} \label{sec:eval_exp}
We evaluate on two standard semantic similarity and relatedness judgement datasets: MEN \citep{bruni2014multimodal} and SimLex-999 \citep{Hill2015}. Model performance is assessed through the Spearman $\rho_s$ rank correlation between the embedding similarity scores for a given pair of words, and human judgements.

For creating multi-modal representations we apply mid-level fusion, concatenating the L2-normalized representations \citep{bruni2014multimodal, Kiela:2016emnlp}, since it allows us to study the information preserved in the individual modalities.

\section{Size and Performance} \label{exp:size_perf}

We perform two experiments.
Firstly, we compare our models while restricting the training data size of $E_L$.
The corpus is sampled randomly to subsets with increasing number of tokens.
In a second experiment, we sub-sample the corpus by token frequency ranges into three equally large parts; HIGH, MEDIUM and LOW range. We sample from the vocabularies of $E_L, E_V$ and $E_S$ by the different frequency ranges coming from $E_L$'s training corpus.
We present these results in App.~\ref{app:size_freq} for we did not find any pattern.

\begin{figure}[t]
\centering
\begin{subfigure}[b]{.49\linewidth}
\includegraphics[width=\textwidth]{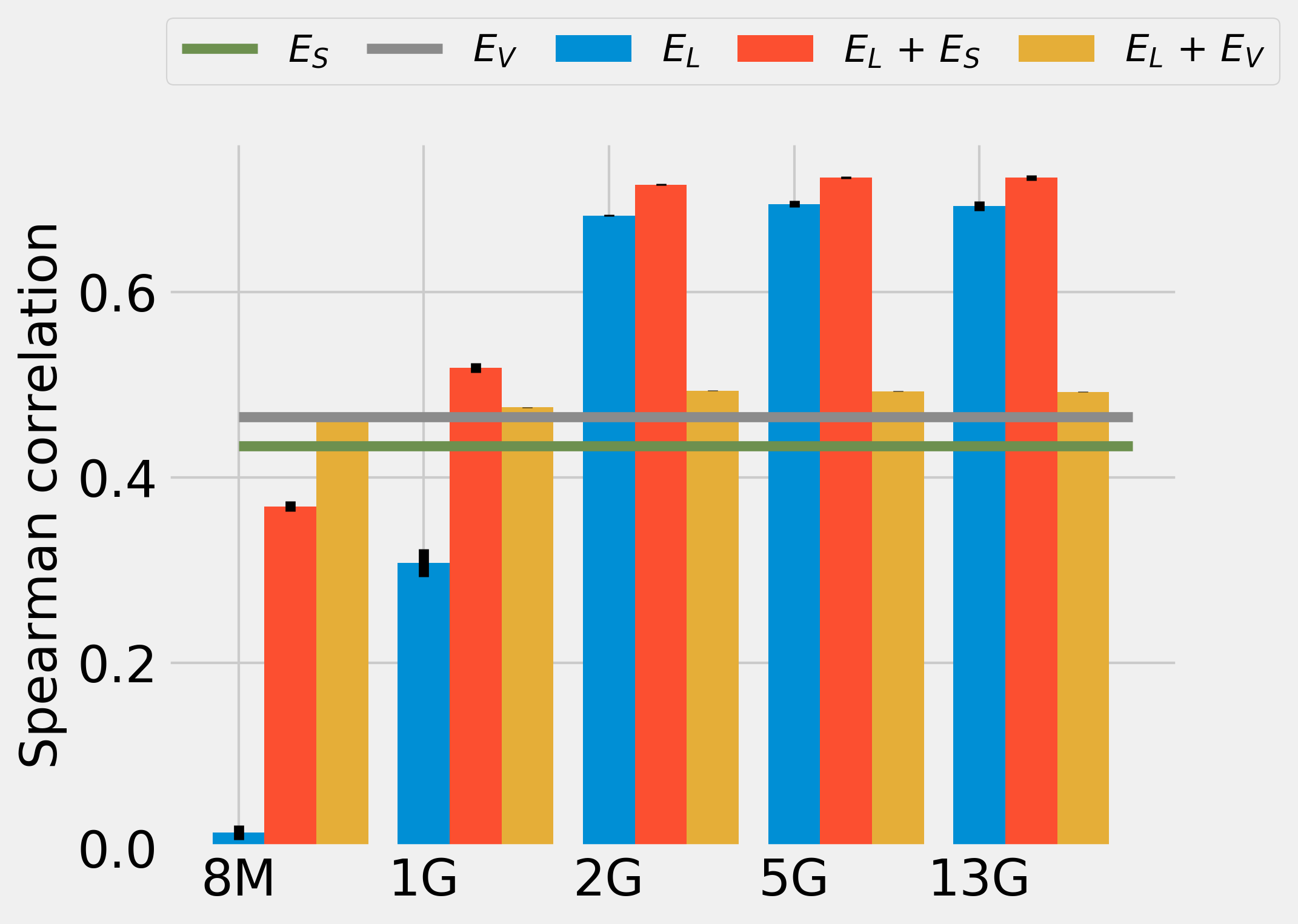}
\caption{MEN}
\label{f:quanities_MEN_common_subs}
\end{subfigure}
\begin{subfigure}[b]{.49\linewidth}
\includegraphics[width=\textwidth]{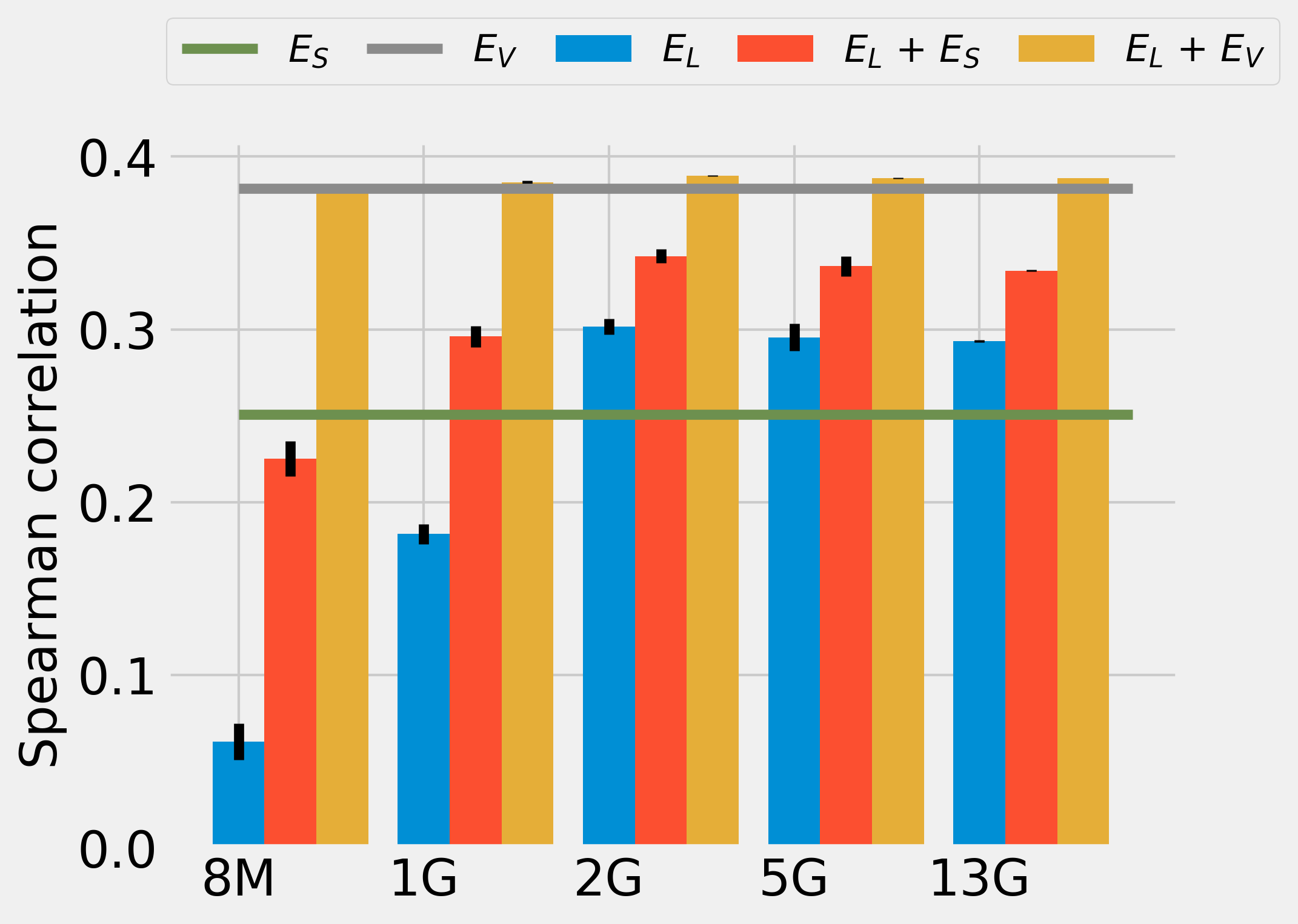}
\caption{SimLex}
\label{f:quanities_SimLex_common_subs}
\end{subfigure}
\caption{Effect of $E_L$ training corpus (token) quantity on performance on the common coverage subsets of evaluation pairs (73\% on MEN, 56\% on SimLex).}
\label{f:quant_common_subs}
\end{figure}

\begin{table}
    \centering
    \setlength\tabcolsep{3pt}
    \begin{tabular}{llllll}
    \hline\hline
    & \textbf{$E_L$} & \textbf{$E_S$} & \textbf{$E_V$} \\\hline
    Model & SGNS & SGNS & ResNet-152 \\
    Tr. Data &  \makecell[tl]{Wikipedia\\2020} & \makecell[tl]{VG\\ScGr} & \makecell[tl]{ImageNet + \\ Google Imgs.}\\
    Size & \makecell[tl]{$\sim$13G\\tokens} & \makecell[tl]{$\sim$9M\\tokens} & \makecell[tl]{$\sim$1.28M + \\15,770\\images (jpg)} \\
    Storage size & 14GB & $\sim$1.8GB & $\sim$140GB \\
    Compr. size & $\sim$5GB & $\sim$0.2GB & $\sim$140GB \\
    \hline
    \end{tabular}
    \caption{Training data sizes.}
    \label{table:data_sizes}
\end{table}

Fig.~\ref{f:quant_common_subs} shows the effect of $E_L$ corpus size on the performance of uni-modal $E_L$ and the combined $E_L + E_S$ and $E_L + E_V$ on the embeddings' common coverage subsets of MEN (Fig.~\ref{f:quanities_MEN_common_subs}) and SimLex (Fig.~\ref{f:quanities_SimLex_common_subs}). $E_S$ and $E_V$ are constant since only $E_L$'s training data is varied. (Results on the full datasets are in App.~\ref{app:size_freq}.) Axis $x$ represents the size of the training corpus (in the number of tokens). Error bars indicate variance after three runs of random down-sampling of the data. Tbl.~\ref{table:data_sizes} gives an account of the amount of training data each model requires. The last line shows the size after compression by Lempel-Ziv coding (LZ77). Since ImageNet images are already in jpg format LZ77 was not able to achieve any further compression.

The first striking result is that $E_S$ alone, with $\sim$9M tokens, outperforms $E_L$, with $\sim$1G tokens, on both evaluation tasks. Secondly, when combined with linguistic data, $E_S$ greatly outperforms $E_V$ on MEN and underperforms it on SimLex, however, their difference becomes marginal as text data increases.
Importantly, $E_S$ achieves this result with orders of magnitude less data than required by $E_V$ (Tbl.~\ref{table:data_sizes}).
Moreover, ResNet-152 with $\sim$6.8G parameters outputs a 1.7 times bigger model (4.8MB) than SGNS, used for $E_L$ and $E_S$ (2.8MB), consisting of 151,200 parameters.\footnote{On the common subset of their vocabularies of 1203 words.}

$E_S$ performs similarly to the FastText VG description model of \citealp{herbelot2020re} on SimLex. The increase of $E_L$ performance is in line with \citealp{sahlgren2016effects} until 2G tokens (they stopped at 1G), after which it plateaus.

Overall, we conclude that our structured visuo-linguistic embedding contributes to a linguistic model in a much more economic way than the image based ones.

\section{Interpretation} \label{exp:interpret}

We aim to peek under the hood of performance results and understand why $E_S$ contributes remarkably despite its small size.

\subsection{Information Gain} \label{exp:MI}

In this experiment we aim to measure the information gain $E_S$ and $E_V$ each contribute when combined with $E_L$. By treating the embedding spaces as samples from multivariate distributions we formulate the question in the following way: \textit{Are two distributions of semantic spaces from different modalities independent from each other?} 

We measure the Mutual Informations: $I(E_L, E_S)$ and $I(E_L, E_V)$. Each embedding space represents samples from a multivariate distribution.
Since $I$ is a special case of divergence, divergence estimators can be employed to estimate it, such as methods based on k-Nearest Neighbor distances ($I_{\!K\!N\!N}$) \citep{wang2009divergence}.
A more robust approach in high dimensions is to introduce non-linearity using a kernel, when calculating the distances, such as the Hilbert-Schmidt Independence Criterion (HSIC) kernel algorithm \citep{gretton2005measuring}.
We applied an open source Python implementation of these algorithms from the Information Theoretical Estimators Toolbox\footnote{\url{https://bitbucket.org/szzoli/ite}} \citep{szabo2014information}.

\begin{figure}
\centering
\begin{subfigure}[b]{.49\linewidth}
\includegraphics[width=\textwidth]{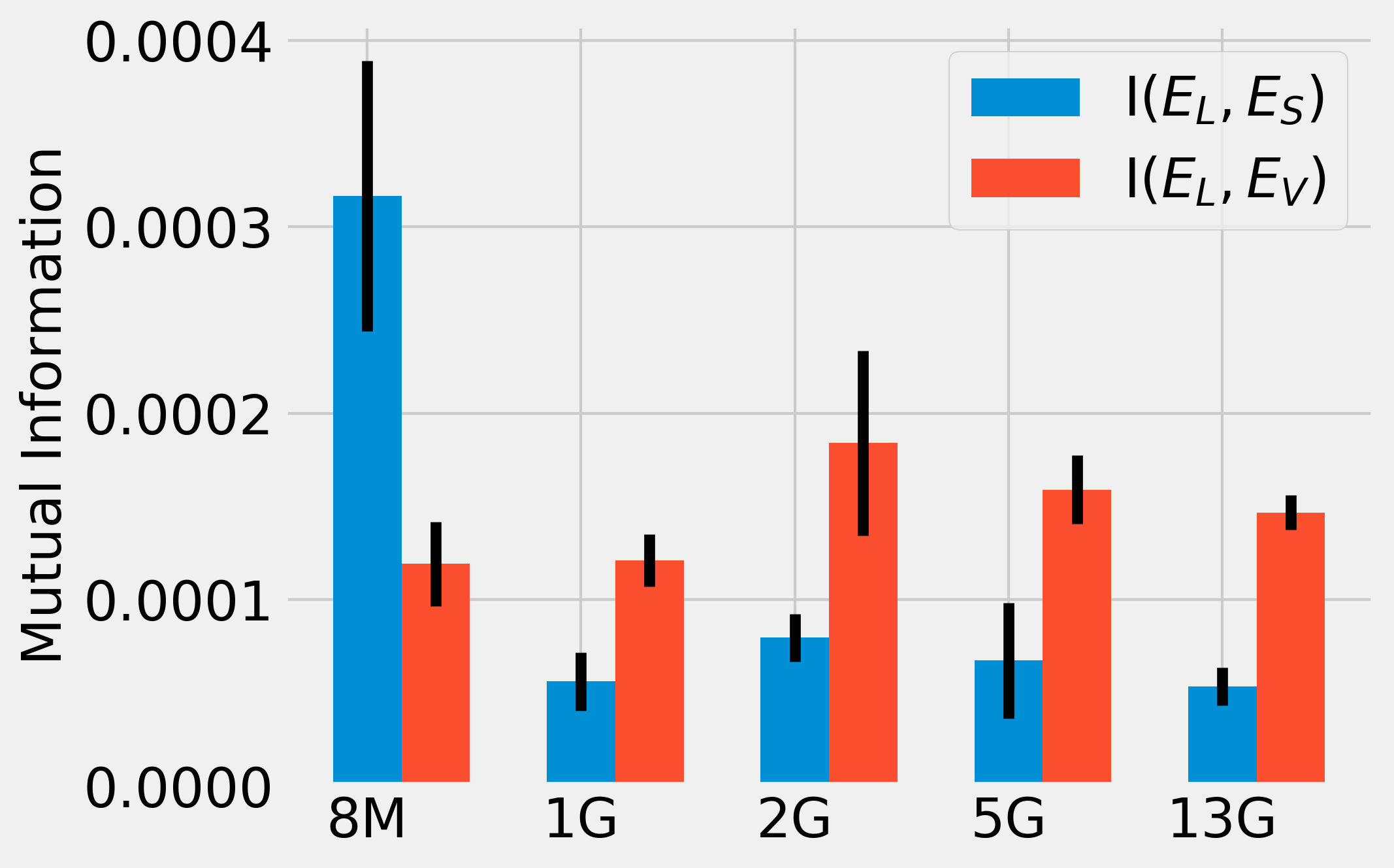}
\caption{$E_L$ quantity ranges}
\label{f:MI_HSIC_sigma-median_quanities}
\end{subfigure}
\begin{subfigure}[b]{.49\linewidth}
\includegraphics[width=\textwidth]{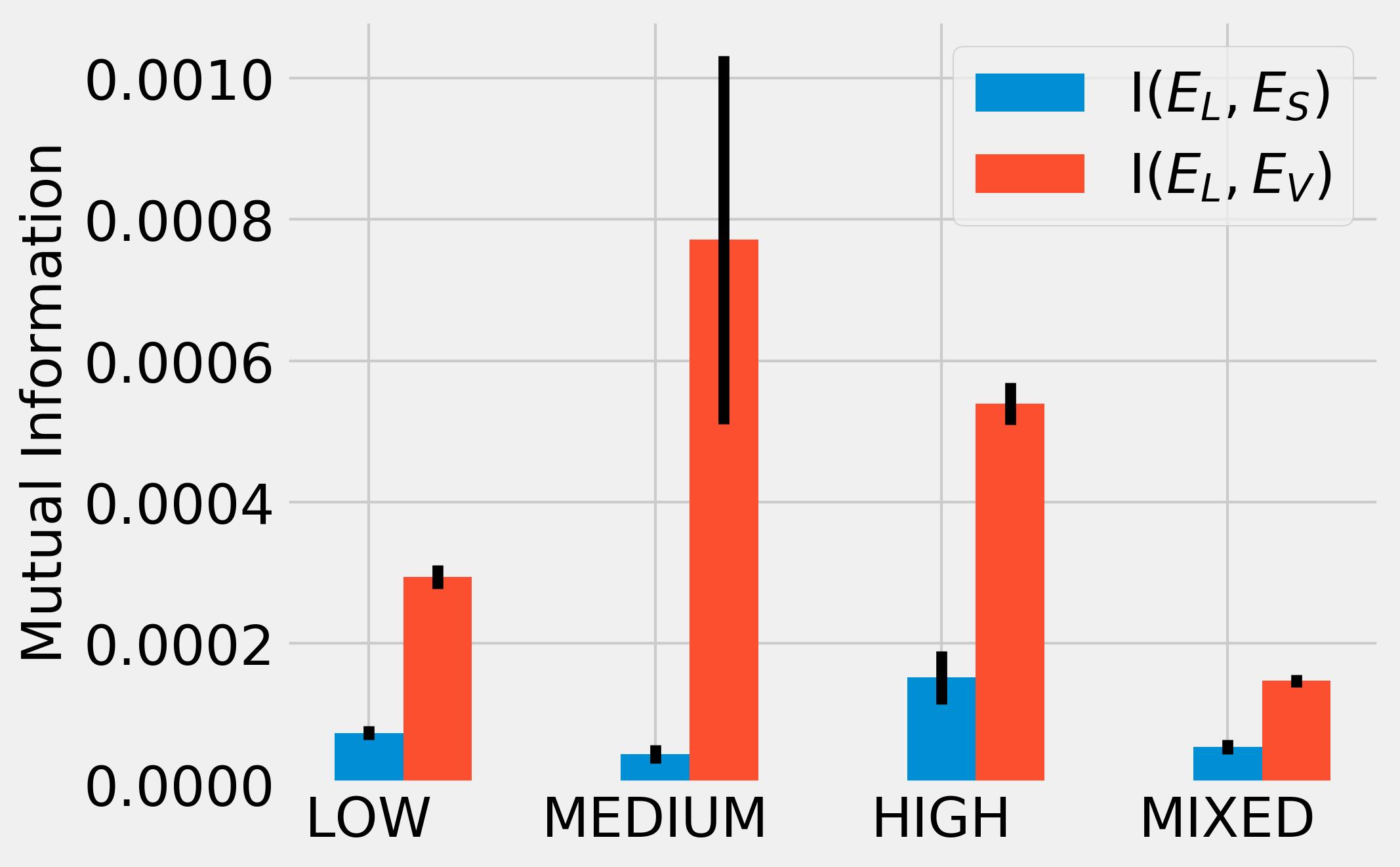}
\caption{$E_L$ frequency ranges}
\label{f:MI_HSIC_sigma-median_freqranges}
\end{subfigure}
\caption{Estimated Mutual Informations: $I(E_L, E_V)$ (red) and $I(E_L, E_S)$ (blue) for different corpus sizes and frequency ranges. Estimated by $I_{\!H\!S\!I\!C}$ algorithm with $\sigma =$ median.}
\label{f:MI}
\end{figure}

In Fig~\ref{f:MI} axis $y$ shows $I(E_L,E_V)$ (red) and $I(E_L,E_S)$ (blue), where $I$ is the estimated Shannon mutual information using the $I_{\!H\!S\!I\!C}$ kernel method. Fig.~\ref{f:MI_HSIC_sigma-median_quanities} presents results across $E_L$ data sizes, Fig.~\ref{f:MI_HSIC_sigma-median_freqranges} includes token frequency ranges (as in Sec.~\ref{exp:size_perf}).
With the exception of the 8M $E_L$ condition, $I(E_L,E_V)$ tend to be greater than $I(E_L,E_S)$, which suggests that $E_S$ is ``more independent'' from the linguistic model $E_L$ than the image based $E_V$. This is surprising, as intuitively the text based $E_S$ is more linguistic than $E_V$, but suggests that $E_S$ provides proportionally  more complementary information to $E_L$ than $E_V$.

To test robustness we ran $I_{\!H\!S\!I\!C}$ for a range of embedding dimensions, which we projected using PCA \citep{wold1987principal}, and tried different $\sigma$ settings. These and the $I_{\!K\!N\!N}$ results support the main findings and are shown in App.~\ref{app:MI}.

\subsection{Qualitative Interpretation of Structure} \label{exp:cluster}

In order to grasp how the concept structures of embeddings differ from each other we investigate their cluster structure.
We perform K-means \citep{macqueen1967some} and Agglomerative \citep{ward1963hierarchical} clustering on them.
To measure the rate of clusterization, when the labels are not known, we use three standard metrics
\textit{Davies–Bouldin Index} (DBI), \textit{Calinski-Harabasz Index} (CHI) and \textit{Silhouette Coefficient} (SC).\footnote{We used their Scikit-learn implementation \url{https://scikit-learn.org/stable/modules/clustering.html\#clustering-performance-evaluation}}
We also present T-SNE visualisations \citep{maaten2008visualizing,wattenberg2016how} of the embeddings.\footnote{We applied the Tensorboard implementation: \url{https://www.tensorflow.org/tensorboard}. We present results for perplexity = 30 as we did not find much difference between the perplexity settings in their suggested range of 5 -- 50.}

\begin{figure}[ht]
	\centering
  \begin{subfigure}[b]{.49\linewidth}
  \includegraphics[width=\textwidth]{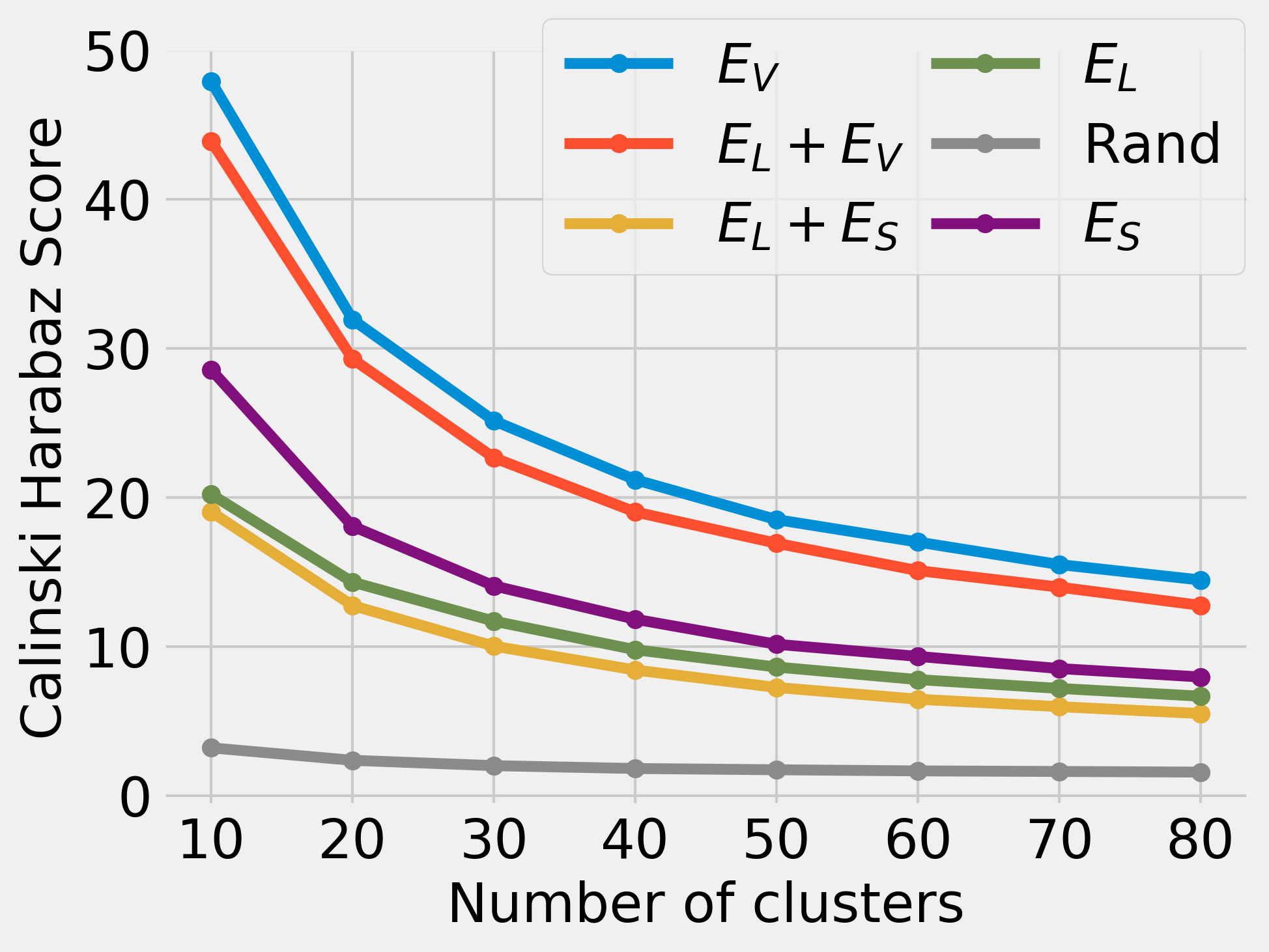}
  \captionsetup{format=hang}
  \caption{Calinski-Harabaz Index. \\ Higher is better.}
  \label{f:Calinski_Harabaz_Score}
  \end{subfigure}
  \begin{subfigure}[b]{.49\linewidth}
  \includegraphics[width=\textwidth]{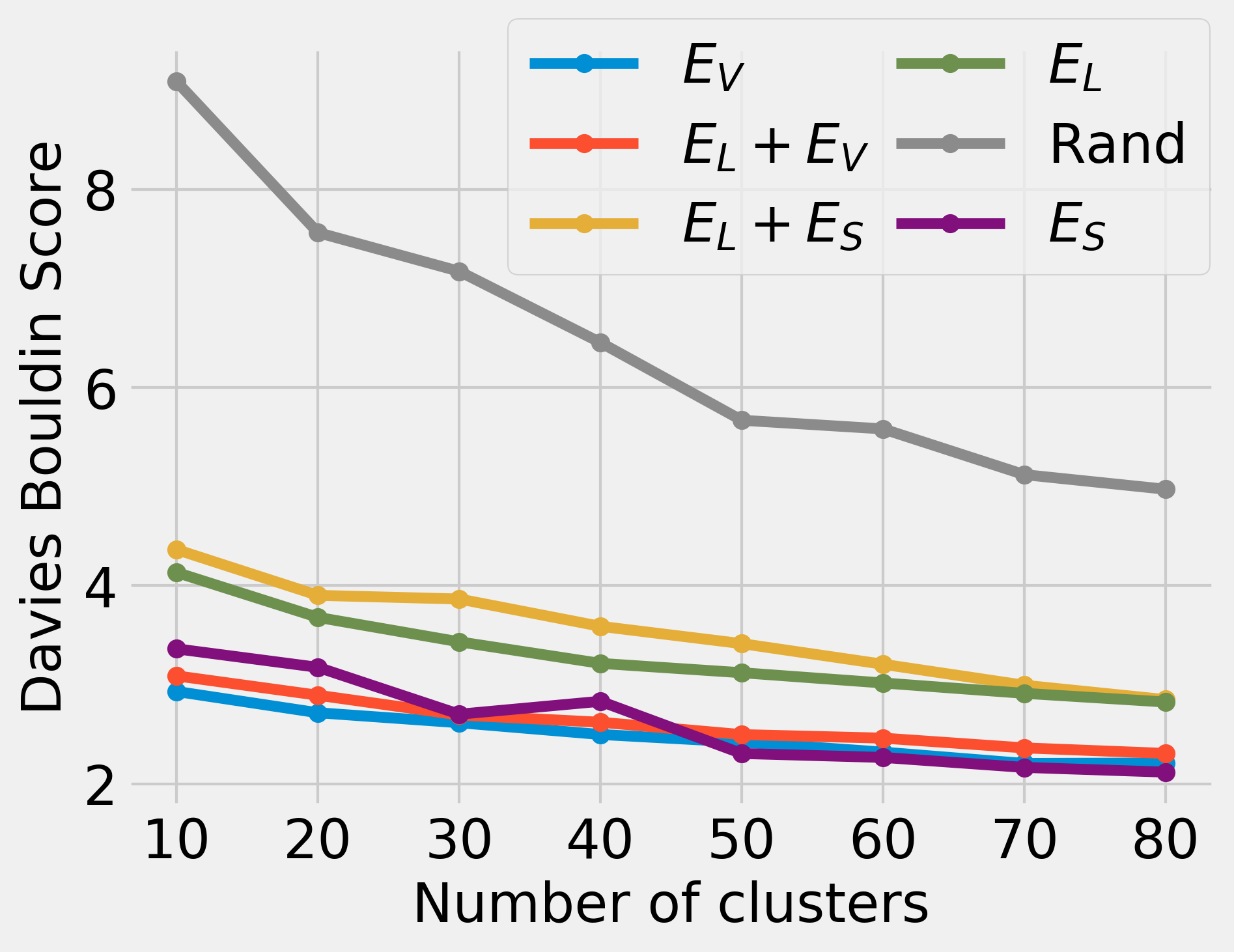}
  \captionsetup{format=hang}
  \caption{Davies-Bouldin Index. \\ Lower is better.}
  \label{f:Davies_Bouldin_Score}
  \end{subfigure}
  \begin{subfigure}[b]{.49\linewidth}
  \includegraphics[width=\textwidth]{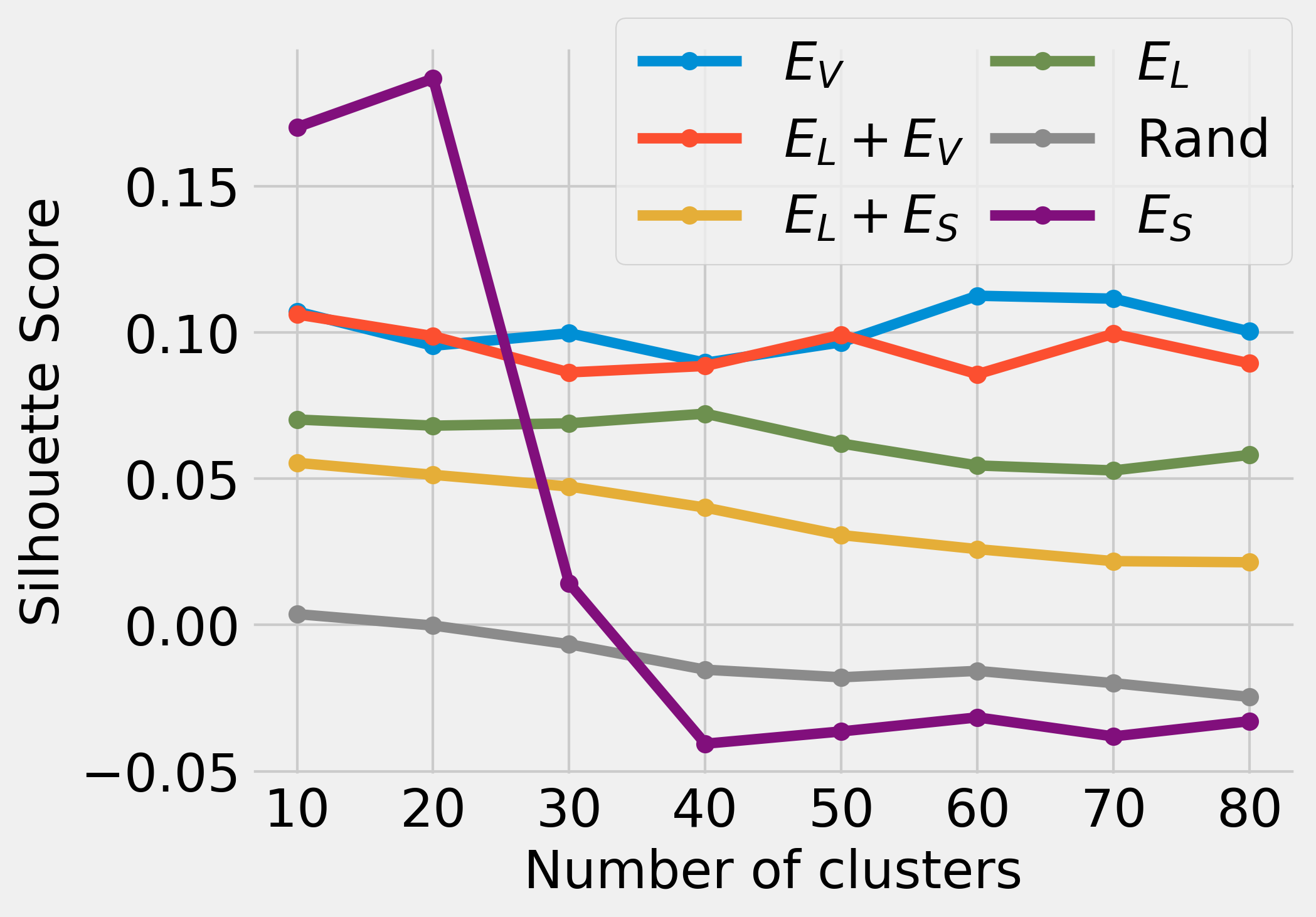}
  \captionsetup{format=hang}
  \caption{Silhouette Coefficient. \\ Higher is better.}
  \label{f:Silhouette_Score}
  \end{subfigure}
	\begin{subfigure}[b]{.49\linewidth}
	\includegraphics[width=\textwidth]{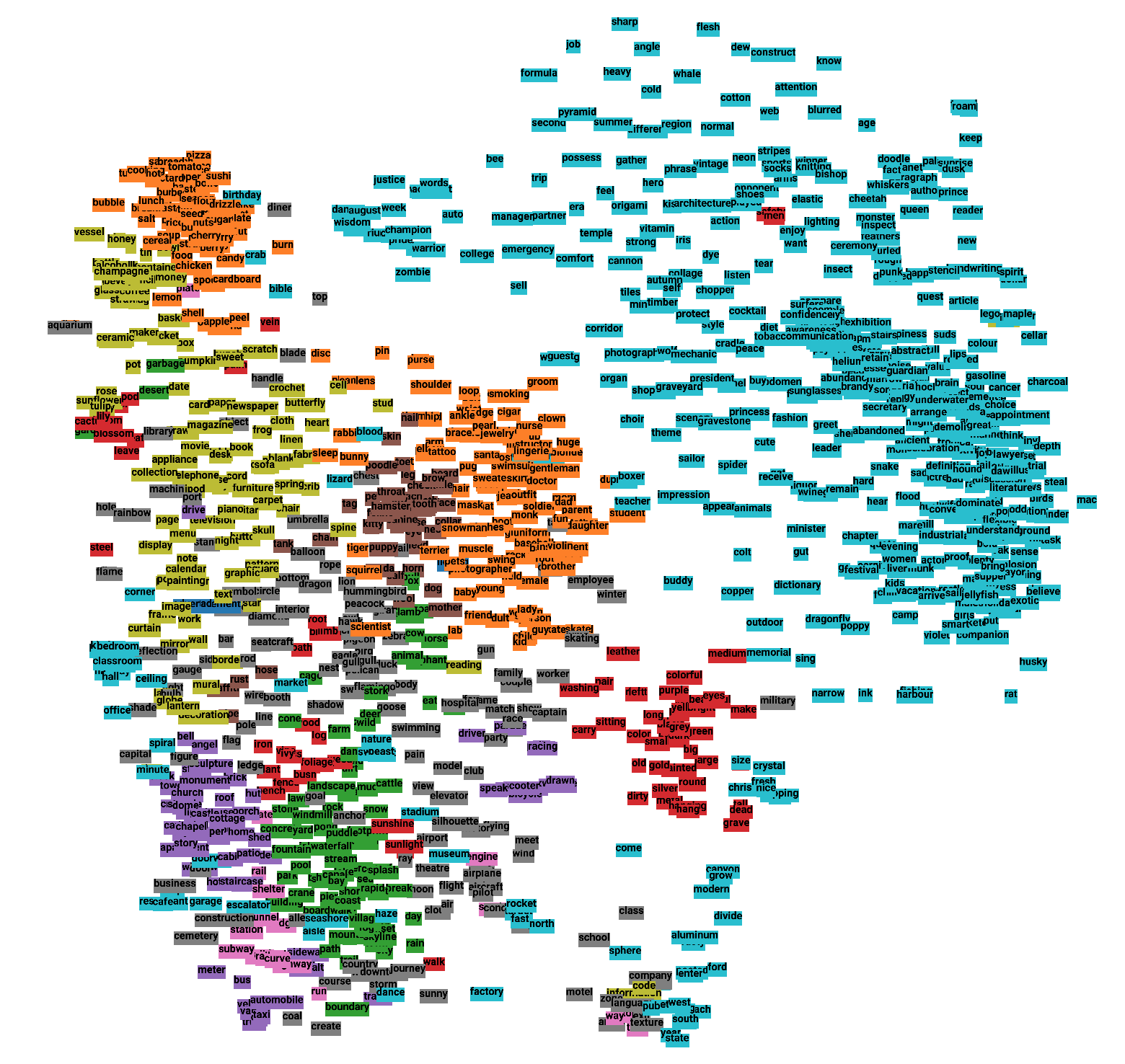}
	\caption{$E_S$}
	\label{f:E_S_20_clusters}
	\end{subfigure}
	\begin{subfigure}[b]{.49\linewidth}
	\includegraphics[width=\textwidth]{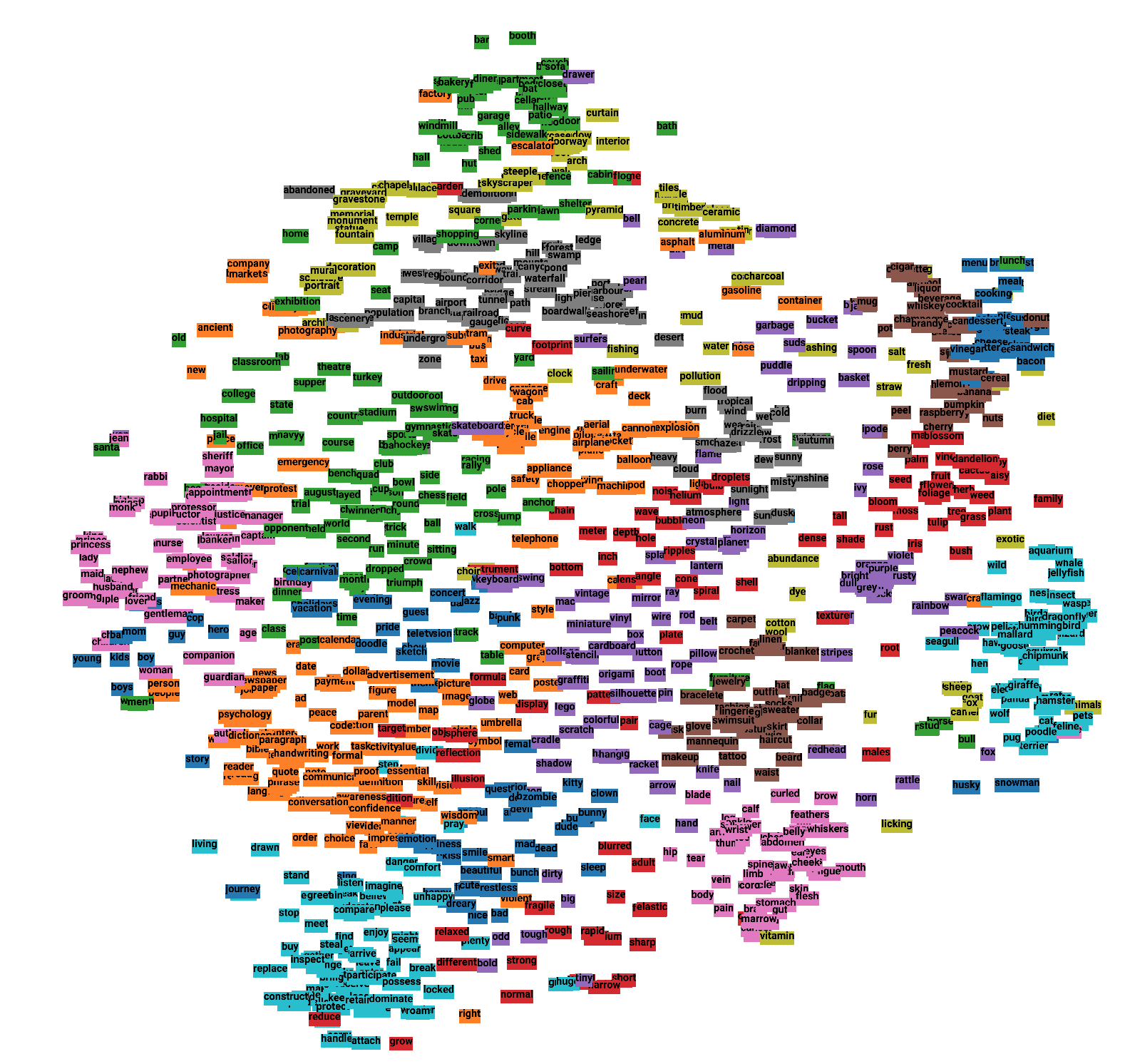}
	\caption{$E_L$}
	\label{f:E_L_20_clusters}
	\end{subfigure}
	\begin{subfigure}[b]{.49\linewidth}
	\includegraphics[width=\textwidth]{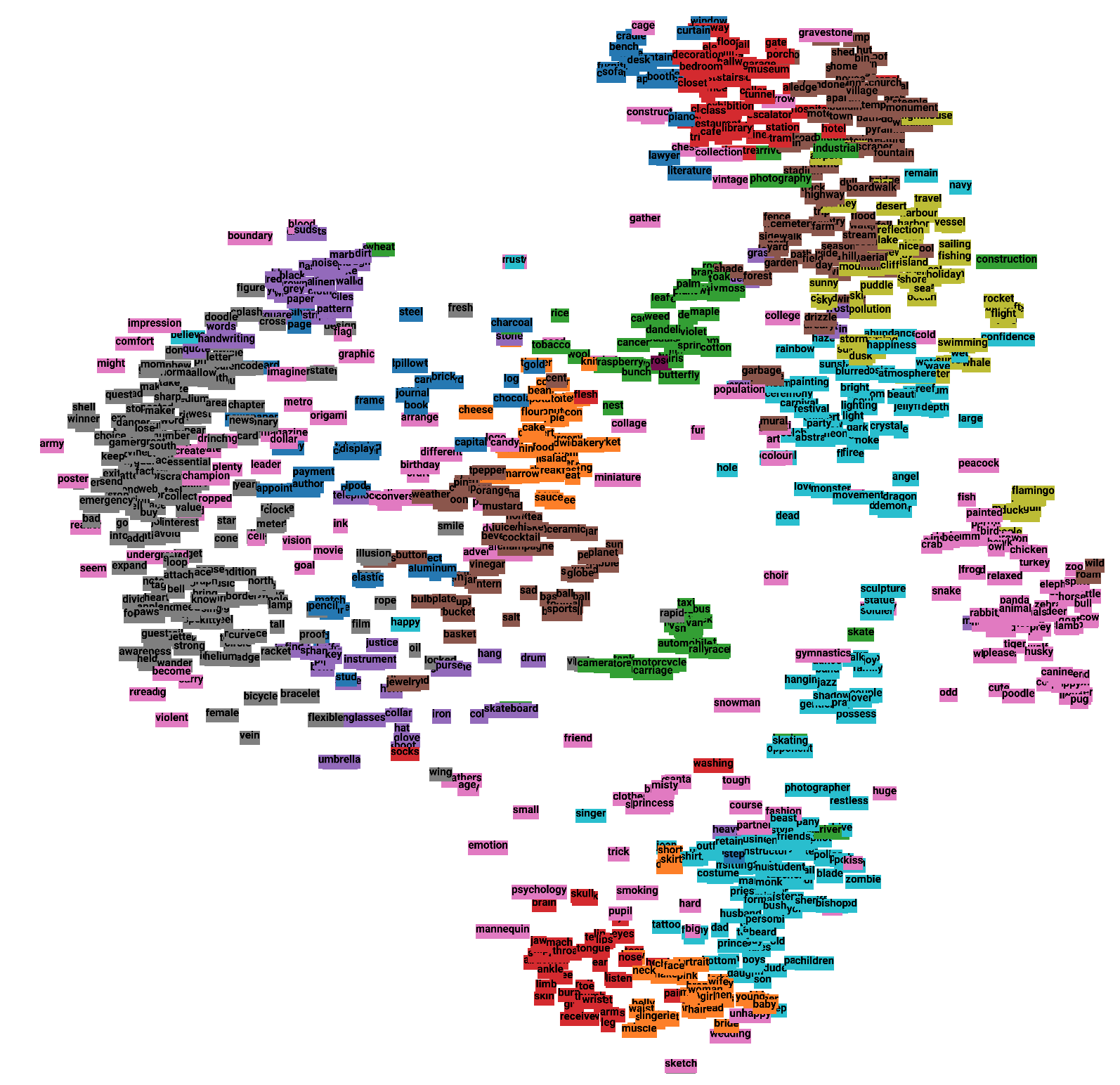}
	\caption{$E_V$}
	\label{f:E_V_20_clusters}
	\end{subfigure}
	\caption{Clustering metrics and T-SNE plots of 20 cluster labels obtained by K-means clustering.}
	\label{f:20_kmeans_clusters}
\end{figure}

Fig.~\ref{f:Calinski_Harabaz_Score} - \ref{f:Silhouette_Score} report on clustering metrics for increasing number of clusters. Fig.~\ref{f:E_S_20_clusters} - \ref{f:E_V_20_clusters} show T-SNE plots of 20 cluster labels obtained by K-means clustering.
On all three metrics, all models outperform the random embedding baseline of the same vocabulary\footnote{The common subset of all embedding vocabularies of 1203 words.} and dimensionality. $E_V$ and $E_L + E_V$ consistently achieve the best scores, however, $E_S$ produces comparable DBI. 
On SC $E_S$ substantially outperforms the others, up until $\sim$20 clusters, when it drops dramatically. This suggests that $E_S$ has a much more cohesive structure of $\sim$20 clusters, but becomes incohesive if we try and break it into more.
This may be an artefact of VG, as \citealp{krishnavisualgenome} reports an average of 17 description clusters on an image.\footnote{Although they clustered averaged pre-trained word representations of region descriptions, not scene graphs, therefore, their results are not directly comparable to ours.}

On the 2D T-SNE projections, $E_V$ appears to have the most defined clusters.
We labelled the 20 cluster case of all embeddings by hand for both K-means and Agglomerative clustering as control.
We also generated labels based on the most common WordNet synsets \citep{miller1995wordnet} occurring in the cluster. We found that $E_S$ clusters represent more concrete concepts and are more visual than $E_L$ but less than $E_V$. Example clusters and a detailed analysis can be found in App.~\ref{app:cluster}.

% \subsubsection{Gamified Data Collection}

% We developed a collaborative gamified data collection app, called \textit{Concept Game}\footnote{\url{http://concept-guessing-game.com/}}, similar to ESP Game \citep{von2004labeling}, but with word lists (clusters) instead of images. We are hoping to collect more human cluster label annotation for different modalities in the future.

\section{Conclusion and Future Work}

We found that our VG Scene-graph based model achieves comparable or better performance in an economic way, by using orders of magnitude less resources than visual models. When combined, it enriches our linguistic model with more divergent information than the image based one. Its clusters represent more concrete concepts, in-between visual and linguistic domains.

Investigating transformers, Bayesian MI estimators and other evaluations could be potential extensions of these studies.
Applying automatically generated scenes graphs would mitigate the main limitation of this approach, which is the manual labour required for creating VG. This would serve as a highly effective tool with important applications for low resource languages.

% \section*{Acknowledgements}

\bibliography{vero_efficient_multimodal}
\bibliographystyle{acl_natbib}

\newpage
\appendix

\section{Comparison of Various Models} \label{app:compare}

\paragraph{Linguistic models} are three FastText \citep{mikolov2018advances}: \textit{wiki-news-300d-1M}, \textit{wiki-news-300d-1M-subword}, \textit{crawl-300d-2M} and a Skip-Gram model (SGNS) trained on a 2013 Wikipedia dump\footnote{\url{https://fasttext.cc/docs/en/english-vectors.html}}.
\paragraph{Visual models} compared: \textit{AlexNet} \citep{Krizhevsky:2012nips}, \textit{ResNet-152} architecture \citep{he2016deep}, VGG \citep{Simonyan:2014arxiv}, trained on Google Images, and two models trained on VG \textit{internal} object bounding box images or on the \textit{whole} images, similarly to \citealp{davis2019deconstructing}.

Results are shown in Tbl.~\ref{t:MEN_full_nopadding}-\ref{t:SimLex_commonsubset_nopadding}.

  \begin{table}[b!]
  \resizebox{\linewidth}{!}{%
  \begin{tabular}{llrr}
  \hline\hline
   \textbf{Models}                      & \textbf{Spearman}             &   \textbf{P-value} &   \textbf{Coverage} \\
  \hline
   \textit{wikinews}                        & 0.7938                        &                  0 &                3000 \\
   \textit{wikinews\_sub}                   & 0.8032                        &                  0 &                3000 \\
   \textit{crawl}                           & \textbf{\color{red}{0.8455}}  &                  0 &                3000 \\
   \textit{SGNS13}                           & 0.6826                        &                  0 &                3000 \\
  \hline
   \textit{Google AlexNet}                  & 0.5025                        &                  0 &                3000 \\
   \textit{Google VGG}                      & 0.5118                        &                  0 &                3000 \\
   \textit{VG-internal}                     & 0.3737                        &                  0 &                2784 \\
   \textit{VG-whole}                        & 0.409                         &                  0 &                2784 \\
   \textit{Google ResNet-152}               & 0.4688                        &                  0 &                3000 \\
  \hline
  \textit{VG SceneGraph}                    & 0.4238                        &                  0 &                2574 \\
  \hline
   \textit{wikinews+Google AlexNet}         & 0.5033                        &                  0 &                3000 \\
   \textit{wikinews+VG SceneGraph}          & 0.6489                        &                  0 &                2574 \\
   \textit{wikinews+Google VGG}             & 0.5129                        &                  0 &                3000 \\
   \textit{wikinews+VG-internal}            & 0.3754                        &                  0 &                2784 \\
   \textit{wikinews+VG-whole}               & 0.4114                        &                  0 &                2784 \\
   \textit{wikinews+Google ResNet-152}      & 0.4768                        &                  0 &                3000 \\
   \textit{wikinews\_sub+Google AlexNet}    & 0.5026                        &                  0 &                3000 \\
   \textit{wikinews\_sub+VG SceneGraph}     & 0.4477                        &                  0 &                2574 \\
   \textit{wikinews\_sub+Google VGG}        & 0.5118                        &                  0 &                3000 \\
   \textit{wikinews\_sub+VG-internal}       & 0.3738                        &                  0 &                2784 \\
   \textit{wikinews\_sub+VG-whole}          & 0.4091                        &                  0 &                2784 \\
   \textit{wikinews\_sub+Google ResNet-152} & 0.4693                        &                  0 &                3000 \\
   \textit{crawl+Google AlexNet}            & 0.5068                        &                  0 &                3000 \\
   \textit{crawl+VG SceneGraph}             & 0.8114                        &                  0 &                2574 \\
   \textit{crawl+Google VGG}                & 0.5176                        &                  0 &                3000 \\
   \textit{crawl+VG-internal}               & 0.3825                        &                  0 &                2784 \\
   \textit{crawl+VG-whole}                  & 0.4213                        &                  0 &                2784 \\
   \textit{crawl+Google ResNet-152}         & 0.5059                        &                  0 &                3000 \\
   \textit{SGNS13+Google AlexNet}            & 0.5046                        &                  0 &                3000 \\
   \textit{SGNS13+VG SceneGraph}             & \textbf{\color{blue}{0.6951}} &                  0 &                2574 \\
   \textit{SGNS13+Google VGG}                & 0.5145                        &                  0 &                3000 \\
   \textit{SGNS13+VG-internal}               & 0.3774                        &                  0 &                2784 \\
   \textit{SGNS13+VG-whole}                  & 0.4148                        &                  0 &                2784 \\
   \textit{SGNS13+Google ResNet-152}         & 0.4846                        &                  0 &                3000 \\
  \hline
  \end{tabular}}
  \caption{Spearman correlation on the full MEN dataset. The table sections contain linguistic, visual and multi-modal embeddings in this order. Red colour signifies the best performance, blue means that the multi-modal embedding outperformed the corresponding uni-modal ones.}
  \label{t:MEN_full_nopadding}
  \end{table}

  \begin{table}[b!]
  \resizebox{\linewidth}{!}{%
  \begin{tabular}{llrr}
  \hline\hline
   \textbf{Models}                      & \textbf{Spearman}             &   \textbf{P-value} &   \textbf{Coverage} \\
  \hline
   \textit{wikinews}                        & 0.45                          &             0      &                 999 \\
   \textit{wikinews\_sub}                   & 0.4409                        &             0      &                 999 \\
   \textit{crawl}                           & \textbf{\color{red}{0.5031}}  &             0      &                 999 \\
   \textit{SGNS13}                           & 0.31                          &             0      &                 999 \\
  \hline
   \textit{Google AlexNet}                  & 0.3401                        &             0      &                 999 \\
   \textit{Google VGG}                      & 0.342                         &             0      &                 999 \\
   \textit{VG-internal}                     & 0.307                         &             0.0016 &                 103 \\
   \textit{VG-whole}                        & 0.1853                        &             0.061  &                 103 \\
   \textit{Google ResNet-152}               & 0.3465                        &             0      &                 999 \\
  \hline
   \textit{VG SceneGraph}                   & 0.2587                        &             0      &                 593 \\
  \hline
   \textit{wikinews+Google AlexNet}         & 0.3406                        &             0      &                 999 \\
   \textit{wikinews+VG SceneGraph}          & 0.3451                        &             0      &                 593 \\
   \textit{wikinews+Google VGG}             & 0.3426                        &             0      &                 999 \\
   \textit{wikinews+VG-internal}            & 0.3072                        &             0.0016 &                 103 \\
   \textit{wikinews+VG-whole}               & 0.1846                        &             0.062  &                 103 \\
   \textit{wikinews+Google ResNet-152}      & 0.3516                        &             0      &                 999 \\
   \textit{wikinews\_sub+Google AlexNet}    & 0.3401                        &             0      &                 999 \\
   \textit{wikinews\_sub+VG SceneGraph}     & 0.2951                        &             0      &                 593 \\
   \textit{wikinews\_sub+Google VGG}        & 0.342                         &             0      &                 999 \\
   \textit{wikinews\_sub+VG-internal}       & 0.307                         &             0.0016 &                 103 \\
   \textit{wikinews\_sub+VG-whole}          & 0.1848                        &             0.0616 &                 103 \\
   \textit{wikinews\_sub+Google ResNet-152} & 0.3471                        &             0      &                 999 \\
   \textit{crawl+Google AlexNet}            & 0.3427                        &             0      &                 999 \\
   \textit{crawl+VG SceneGraph}             & 0.4438                        &             0      &                 593 \\
   \textit{crawl+Google VGG}                & 0.3448                        &             0      &                 999 \\
   \textit{crawl+VG-internal}               & 0.3122                        &             0.0013 &                 103 \\
   \textit{crawl+VG-whole}                  & 0.1891                        &             0.0557 &                 103 \\
   \textit{crawl+Google ResNet-152}         & 0.3687                        &             0      &                 999 \\
   \textit{SGNS13+Google AlexNet}            & \textbf{\color{blue}{0.341}}  &             0      &                 999 \\
   \textit{SGNS13+VG SceneGraph}             & 0.2914                        &             0      &                 593 \\
   \textit{SGNS13+Google VGG}                & \textbf{\color{blue}{0.3428}} &             0      &                 999 \\
   \textit{SGNS13+VG-internal}               & 0.3076                        &             0.0016 &                 103 \\
   \textit{SGNS13+VG-whole}                  & 0.1832                        &             0.064  &                 103 \\
   \textit{SGNS13+Google ResNet-152}         & \textbf{\color{blue}{0.3539}} &             0      &                 999 \\
  \hline
  \end{tabular}}
  \caption{Spearman correlation on the full SimLex dataset. The table sections contain linguistic, visual and multi-modal embeddings in this order. Red colour signifies the best performance, blue means that the multi-modal embedding outperformed the corresponding uni-modal ones.}
  \label{t:SimLex_full_nopadding}
  \end{table}

  \begin{table}[b!]
  \resizebox{\linewidth}{!}{%
  \begin{tabular}{llrr}
  \hline\hline
   \textbf{Models}                      & \textbf{Spearman}            &   \textbf{P-value} &   \textbf{Coverage} \\
  \hline
   \textit{wikinews}                        & 0.7989                       &                  0 &                2481 \\
   \textit{wikinews\_sub}                   & 0.8035                       &                  0 &                2481 \\
   \textit{crawl}                           & \textbf{\color{red}{0.8434}} &                  0 &                2481 \\
   \textit{SGNS13}                           & 0.673                        &                  0 &                2481 \\
  \hline
   \textit{Google AlexNet}                  & 0.5151                       &                  0 &                2481 \\
   \textit{Google VGG}                      & 0.5146                       &                  0 &                2481 \\
   \textit{VG-internal}                     & 0.3751                       &                  0 &                2481 \\
   \textit{VG-whole}                        & 0.4057                       &                  0 &                2481 \\
   \textit{Google ResNet-152}               & 0.4742                       &                  0 &                2481 \\
  \hline
   \textit{VG SceneGraph}                   & 0.4421                       &                  0 &                2481 \\
  \hline
   \textit{wikinews+Google AlexNet}         & 0.5159                       &                  0 &                2481 \\
   \textit{wikinews+VG SceneGraph}          & 0.6561                       &                  0 &                2481 \\
   \textit{wikinews+Google VGG}             & 0.5158                       &                  0 &                2481 \\
   \textit{wikinews+VG-internal}            & 0.3769                       &                  0 &                2481 \\
   \textit{wikinews+VG-whole}               & 0.4083                       &                  0 &                2481 \\
   \textit{wikinews+Google ResNet-152}      & 0.482                        &                  0 &                2481 \\
   \textit{wikinews\_sub+Google AlexNet}    & 0.5151                       &                  0 &                2481 \\
   \textit{wikinews\_sub+VG SceneGraph}     & 0.4622                       &                  0 &                2481 \\
   \textit{wikinews\_sub+Google VGG}        & 0.5147                       &                  0 &                2481 \\
   \textit{wikinews\_sub+VG-internal}       & 0.3752                       &                  0 &                2481 \\
   \textit{wikinews\_sub+VG-whole}          & 0.4058                       &                  0 &                2481 \\
   \textit{wikinews\_sub+Google ResNet-152} & 0.4747                       &                  0 &                2481 \\
   \textit{crawl+Google AlexNet}            & 0.5191                       &                  0 &                2481 \\
   \textit{crawl+VG SceneGraph}             & 0.8144                       &                  0 &                2481 \\
   \textit{crawl+Google VGG}                & 0.5204                       &                  0 &                2481 \\
   \textit{crawl+VG-internal}               & 0.3841                       &                  0 &                2481 \\
   \textit{crawl+VG-whole}                  & 0.4187                       &                  0 &                2481 \\
   \textit{crawl+Google ResNet-152}         & 0.5105                       &                  0 &                2481 \\
   \textit{SGNS13+Google AlexNet}            & 0.5169                       &                  0 &                2481 \\
   \textit{SGNS13+VG SceneGraph}             & \textbf{\color{blue}{0.697}} &                  0 &                2481 \\
   \textit{SGNS13+Google VGG}                & 0.5172                       &                  0 &                2481 \\
   \textit{SGNS13+VG-internal}               & 0.3789                       &                  0 &                2481 \\
   \textit{SGNS13+VG-whole}                  & 0.412                        &                  0 &                2481 \\
   \textit{SGNS13+Google ResNet-152}         & 0.4887                       &                  0 &                2481 \\
  \hline
  \end{tabular}}
  \caption{Spearman correlation on the common subset of the MEN dataset. The table sections contain linguistic, visual and multi-modal embeddings in this order. Red colour signifies the best performance, blue means that the multi-modal embedding outperformed the corresponding uni-modal ones.}
  \label{t:MEN_commonsubset_nopadding}
  \end{table}

  \begin{table}[b!]
  \resizebox{\linewidth}{!}{%
  \begin{tabular}{llrr}
  \hline\hline
   \textbf{Models}                      & \textbf{Spearman}             &   \textbf{P-value} &   \textbf{Coverage} \\
  \hline
   \textit{wikinews}                        & 0.2818                        &             0.0039 &                 103 \\
   \textit{wikinews\_sub}                   & 0.2454                        &             0.0125 &                 103 \\
   \textit{crawl}                           & 0.3748                        &             0.0001 &                 103 \\
   \textit{SGNS13}                           & 0.1146                        &             0.2491 &                 103 \\
  \hline
   \textit{Google AlexNet}                  & \textbf{\color{red}{0.5539}}  &             0      &                 103 \\
   \textit{Google VGG}                      & 0.5261                        &             0      &                 103 \\
   \textit{VG-internal}                     & 0.307                         &             0.0016 &                 103 \\
   \textit{VG-whole}                        & 0.1853                        &             0.061  &                 103 \\
   \textit{Google ResNet-152}               & 0.5015                        &             0      &                 103 \\
  \hline
   \textit{VG SceneGraph}                   & 0.2992                        &             0.0021 &                 103 \\
  \hline
   \textit{wikinews+Google AlexNet}         & \textbf{\color{red}{0.5539}}  &             0      &                 103 \\
   \textit{wikinews+VG SceneGraph}          & 0.2919                        &             0.0028 &                 103 \\
   \textit{wikinews+Google VGG}             & \textbf{\color{blue}{0.5261}} &             0      &                 103 \\
   \textit{wikinews+VG-internal}            & \textbf{\color{blue}{0.3072}} &             0.0016 &                 103 \\
   \textit{wikinews+VG-whole}               & 0.1846                        &             0.062  &                 103 \\
   \textit{wikinews+Google ResNet-152}      & 0.5003                        &             0      &                 103 \\
   \textit{wikinews\_sub+Google AlexNet}    & \textbf{\color{red}{0.5539}}  &             0      &                 103 \\
   \textit{wikinews\_sub+VG SceneGraph}     & 0.2979                        &             0.0022 &                 103 \\
   \textit{wikinews\_sub+Google VGG}        & 0.5261                        &             0      &                 103 \\
   \textit{wikinews\_sub+VG-internal}       & 0.307                         &             0.0016 &                 103 \\
   \textit{wikinews\_sub+VG-whole}          & 0.1848                        &             0.0616 &                 103 \\
   \textit{wikinews\_sub+Google ResNet-152} & 0.5006                        &             0      &                 103 \\
   \textit{crawl+Google AlexNet}            & 0.5518                        &             0      &                 103 \\
   \textit{crawl+VG SceneGraph}             & 0.341                         &             0.0004 &                 103 \\
   \textit{crawl+Google VGG}                & 0.5247                        &             0      &                 103 \\
   \textit{crawl+VG-internal}               & 0.3122                        &             0.0013 &                 103 \\
   \textit{crawl+VG-whole}                  & 0.1891                        &             0.0557 &                 103 \\
   \textit{crawl+Google ResNet-152}         & 0.4893                        &             0      &                 103 \\
   \textit{SGNS13+Google AlexNet}            & 0.5517                        &             0      &                 103 \\
   \textit{SGNS13+VG SceneGraph}             & 0.25                          &             0.0109 &                 103 \\
   \textit{SGNS13+Google VGG}                & \textbf{\color{blue}{0.5264}} &             0      &                 103 \\
   \textit{SGNS13+VG-internal}               & \textbf{\color{blue}{0.3076}} &             0.0016 &                 103 \\
   \textit{SGNS13+VG-whole}                  & 0.1832                        &             0.064  &                 103 \\
   \textit{SGNS13+Google ResNet-152}         & 0.4922                        &             0      &                 103 \\
  \hline
  \end{tabular}}
  \caption{Spearman correlation on the common subset of the SimLex dataset. The table sections contain linguistic, visual and multi-modal embeddings in this order. Red colour signifies the best performance, blue means that the multi-modal embedding outperformed the corresponding uni-modal ones.}
  \label{t:SimLex_commonsubset_nopadding}
  \end{table}

\section{Size, Frequency and Performance} \label{app:size_freq}

\begin{figure}[h!]
\centering
\begin{subfigure}[b]{.49\linewidth}
\includegraphics[width=\textwidth]{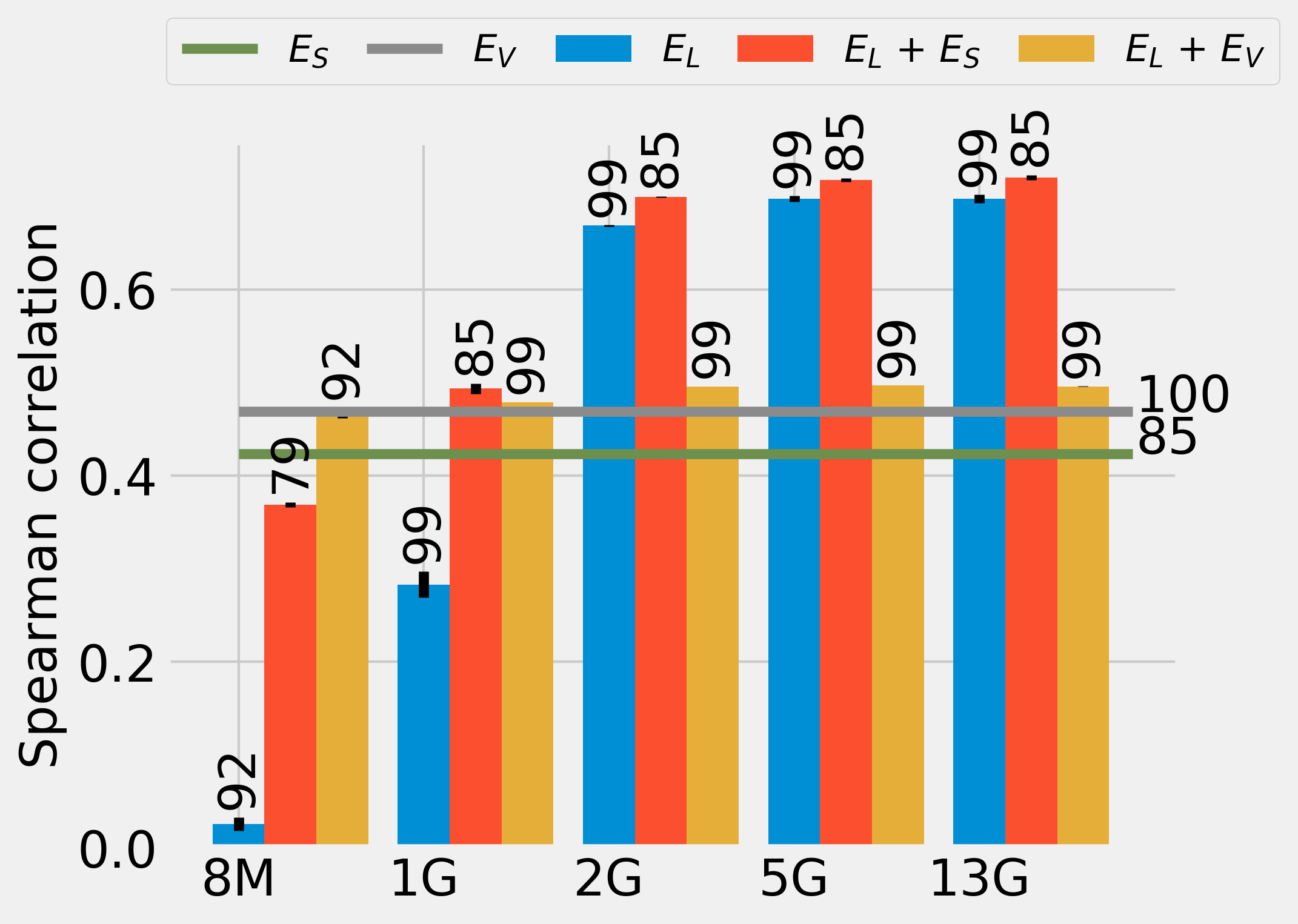}
\caption{MEN, quantity}
\label{f:quanities_MEN}
\end{subfigure}
\begin{subfigure}[b]{.49\linewidth}
\includegraphics[width=\textwidth]{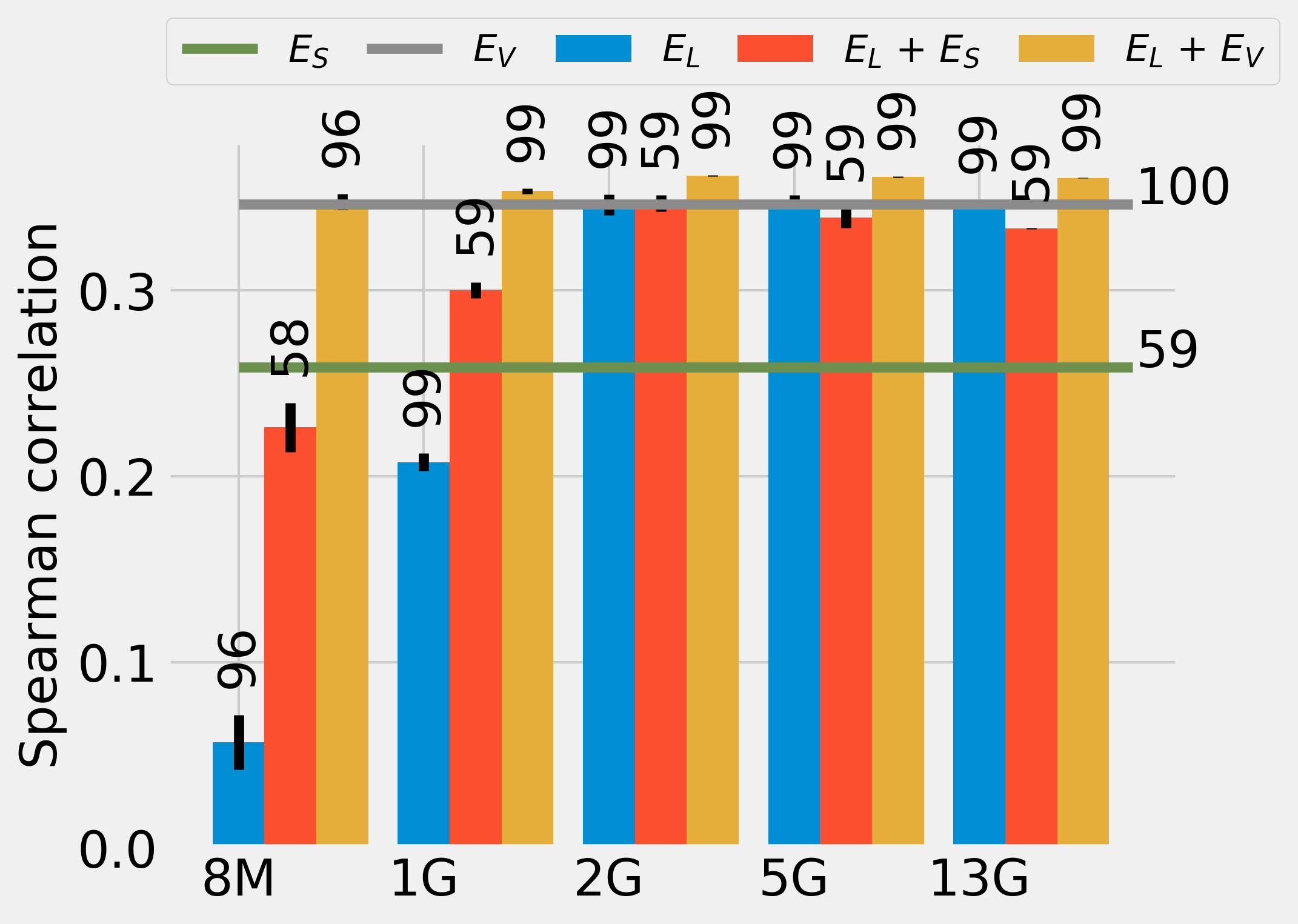}
\caption{SimLex, quantity}
\label{f:quanities_SimLex}
\end{subfigure}
\begin{subfigure}[b]{.49\linewidth}
\includegraphics[width=\textwidth]{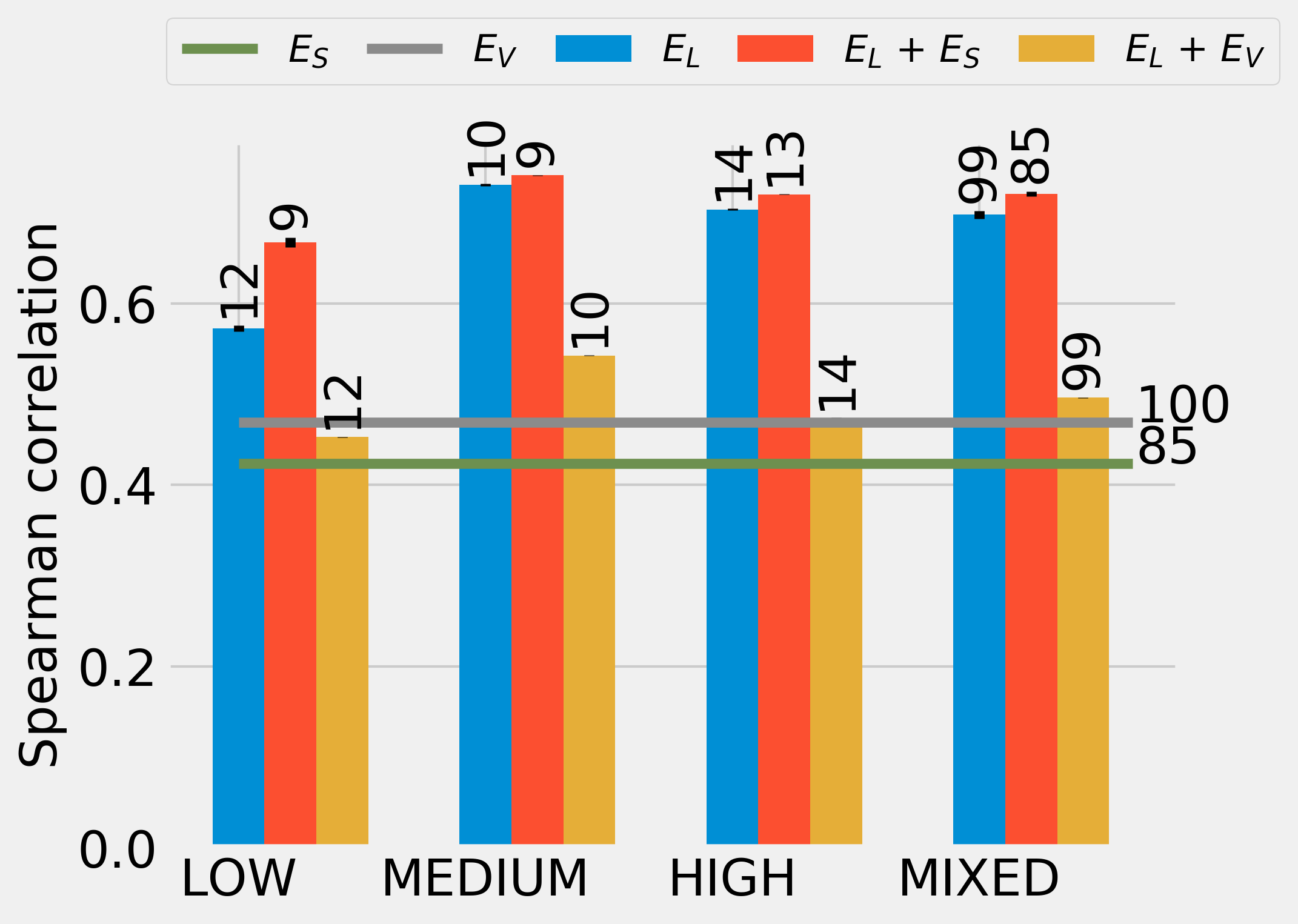}
\caption{MEN, frequency}
\label{f:eqsplit_freqranges_MEN}
\end{subfigure}
\begin{subfigure}[b]{.49\linewidth}
\includegraphics[width=\textwidth]{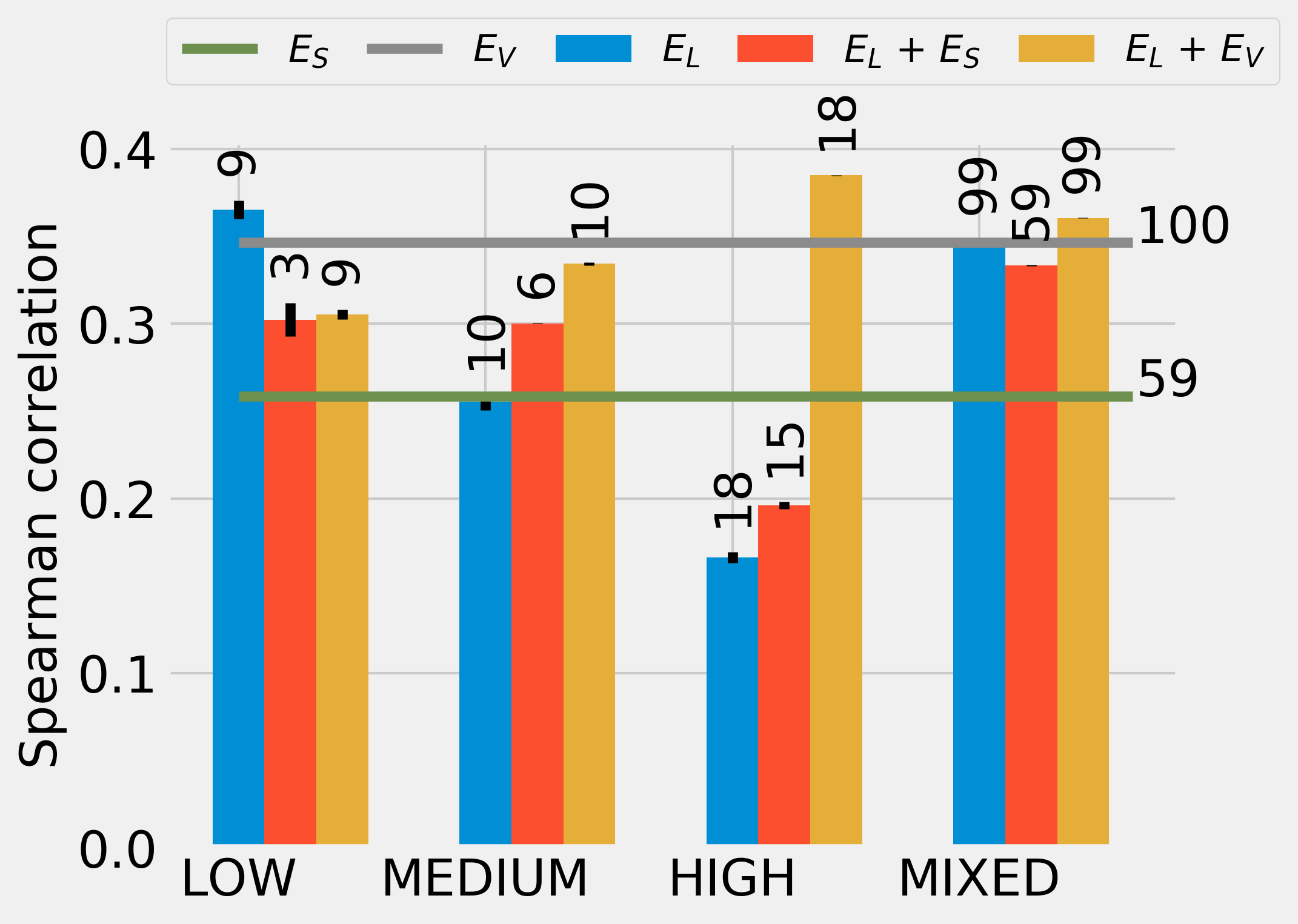}
\caption{SimLex, frequency}
\label{f:eqsplit_freqranges_SimLex}
\end{subfigure}
\caption{Effect of $E_L$ training corpus quantity and word frequency on performance. Numbers on top of the bars and on the lines indicate the coverage of evaluation dataset pairs (where both words are in the embedding vocabulary) in percentages.}
\label{f:fquant_freq_full}
\end{figure}

Fig.~\ref{f:fquant_freq_full} presents performance results controlled for $E_L$ data size and token frequency ranges on the full evaluation datasets.

\section{Information Gain} \label{app:MI}

\begin{figure}[t!]
\centering
\begin{subfigure}[t]{.49\linewidth}
\includegraphics[width=\textwidth]{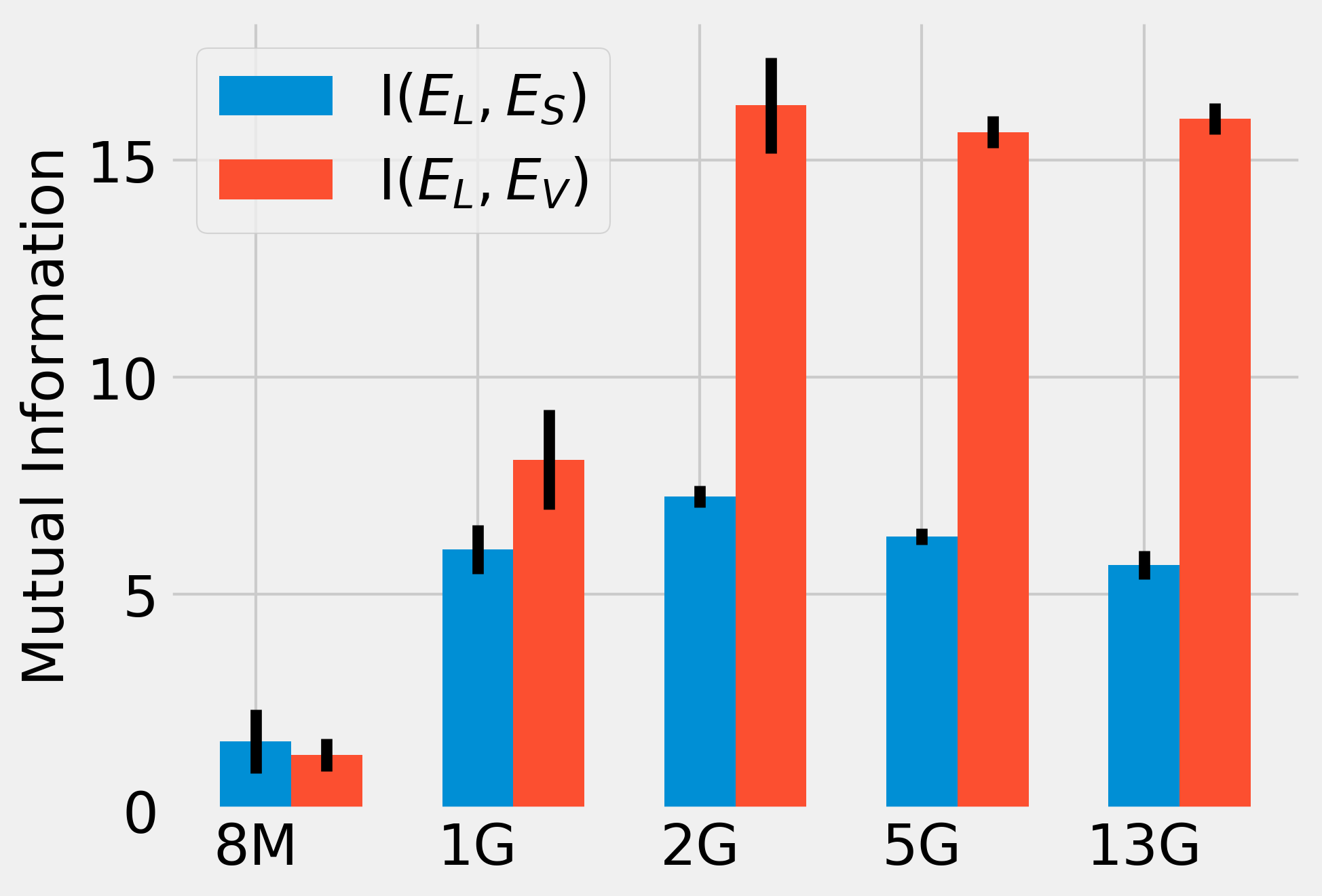}
\caption{$I_{\!K\!N\!N}$}
\label{f:MI_knn_quanities}
\end{subfigure}
\begin{subfigure}[t]{.49\linewidth}
\includegraphics[width=\textwidth]{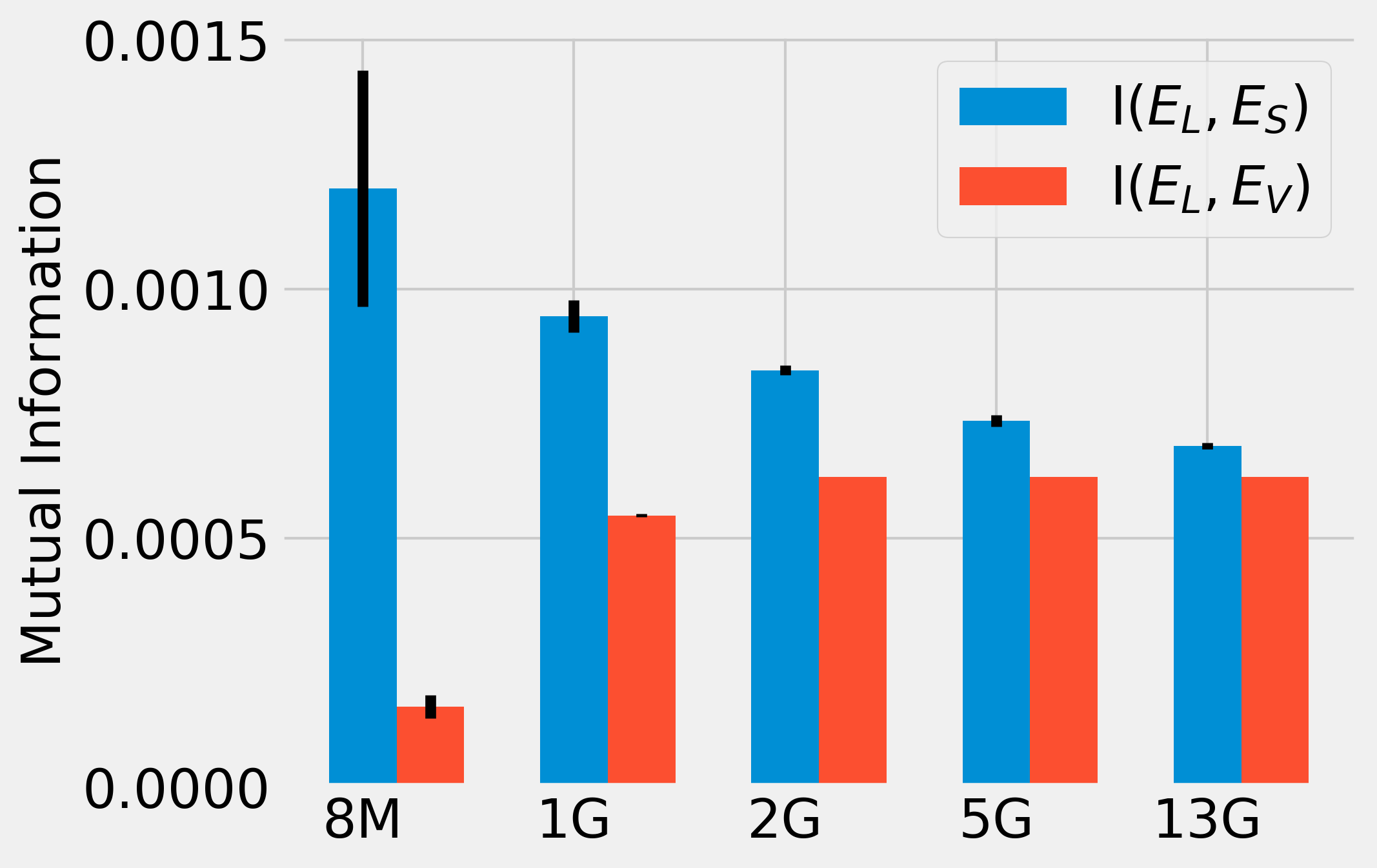}
\caption{$I_{\!H\!S\!I\!C}$, $\sigma = 1$, $d = max$}
\label{f:MI_HSIC_sigma1_quanities}
\end{subfigure}
\begin{subfigure}[t]{.49\linewidth}
\includegraphics[width=\textwidth]{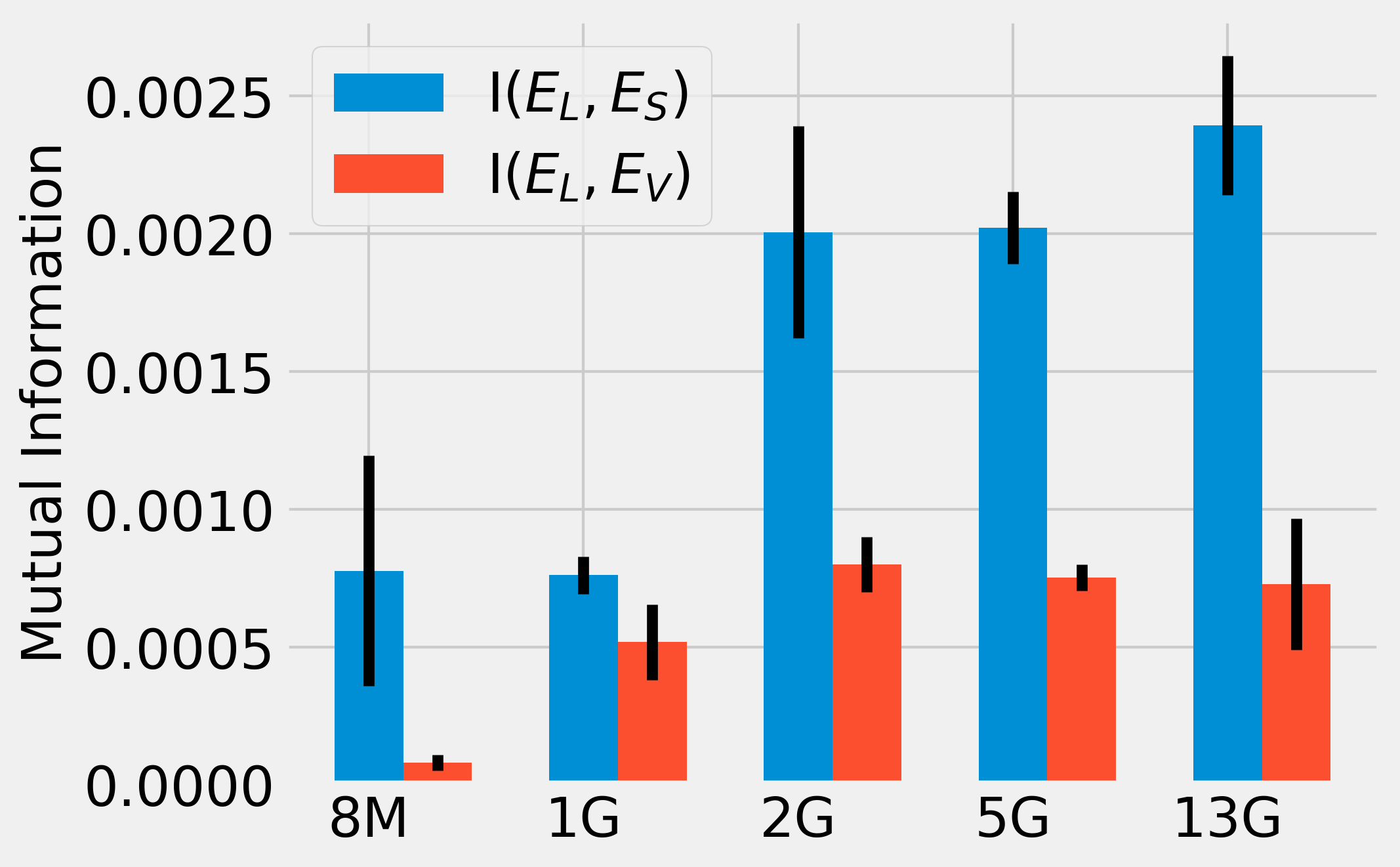}
\caption{$I_{\!H\!S\!I\!C}$, $\sigma$: median, $d = 3$}
\label{f:MI_HSIC_sigma-median_d3_quanities}
\end{subfigure}
\begin{subfigure}[t]{.49\linewidth}
\includegraphics[width=\textwidth]{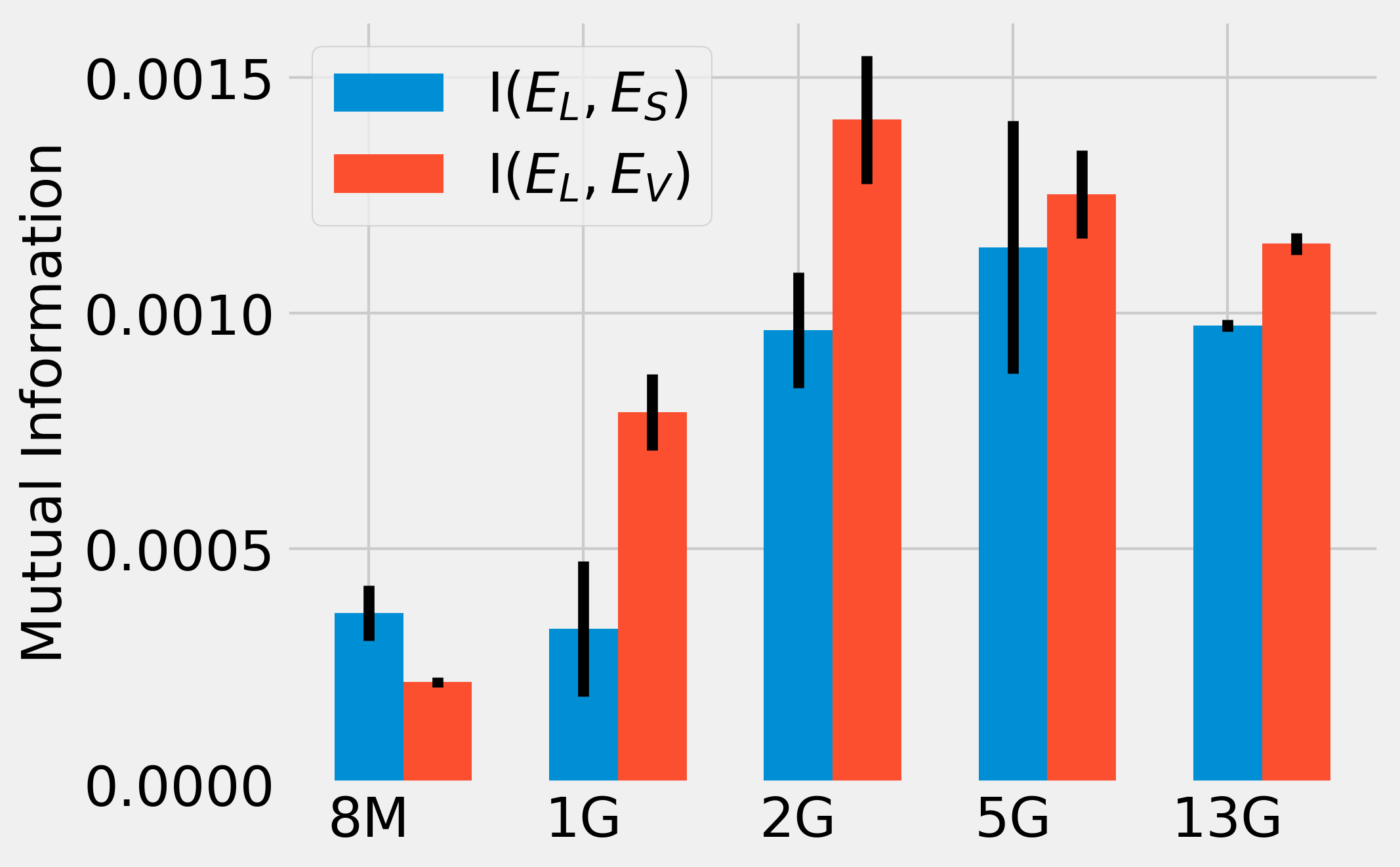}
\caption{$I_{\!H\!S\!I\!C}$, $\sigma$: median, $d = 10$}
\label{f:MI_HSIC_sigma-median_d10_quanities}
\end{subfigure}
\begin{subfigure}[t]{.49\linewidth}
\includegraphics[width=\textwidth]{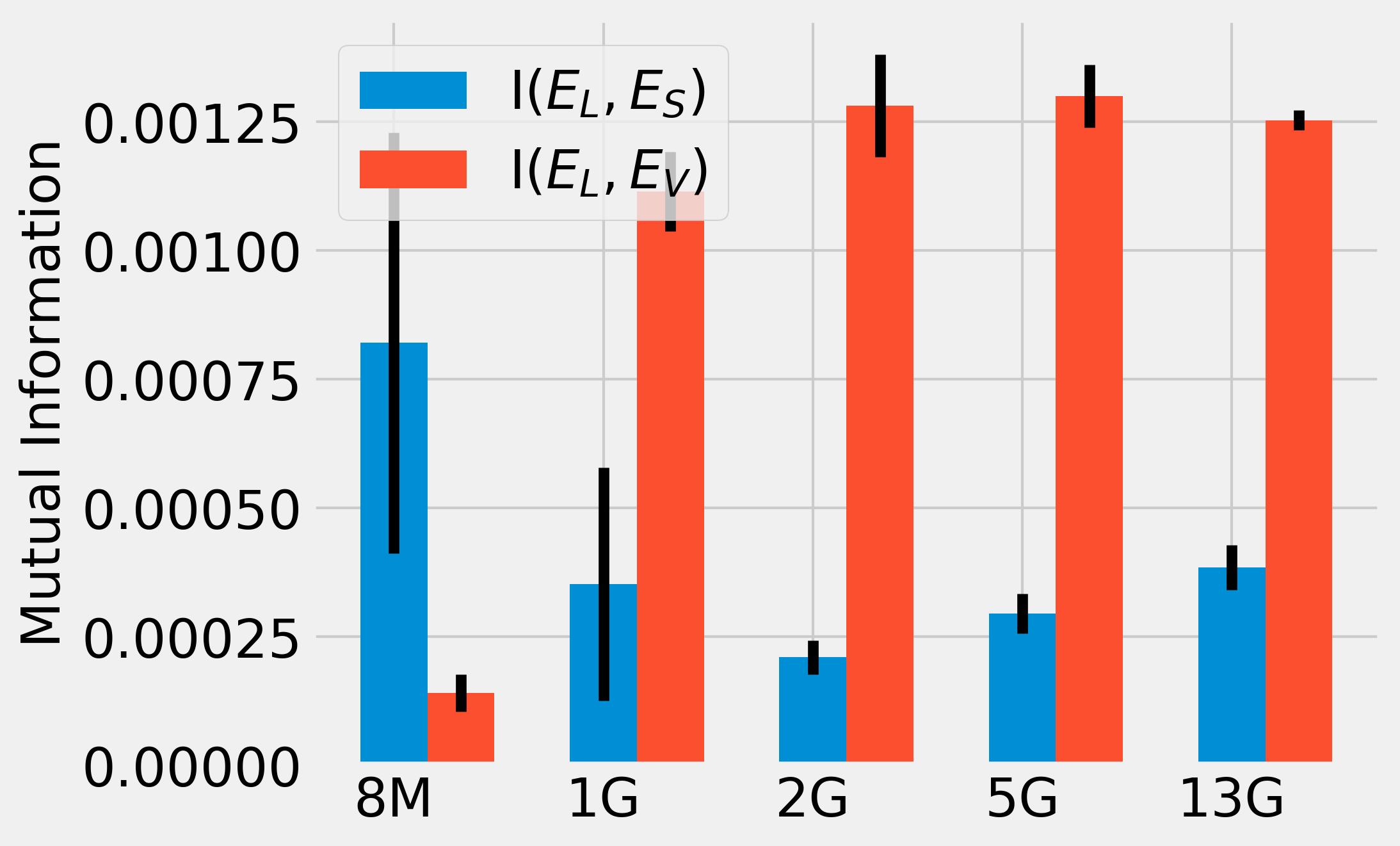}
\caption{$I_{\!H\!S\!I\!C}$, $\sigma$: median, $d = 13$}
\label{f:MI_HSIC_sigma-median_d13_quanities}
\end{subfigure}
\begin{subfigure}[t]{.49\linewidth}
\includegraphics[width=\textwidth]{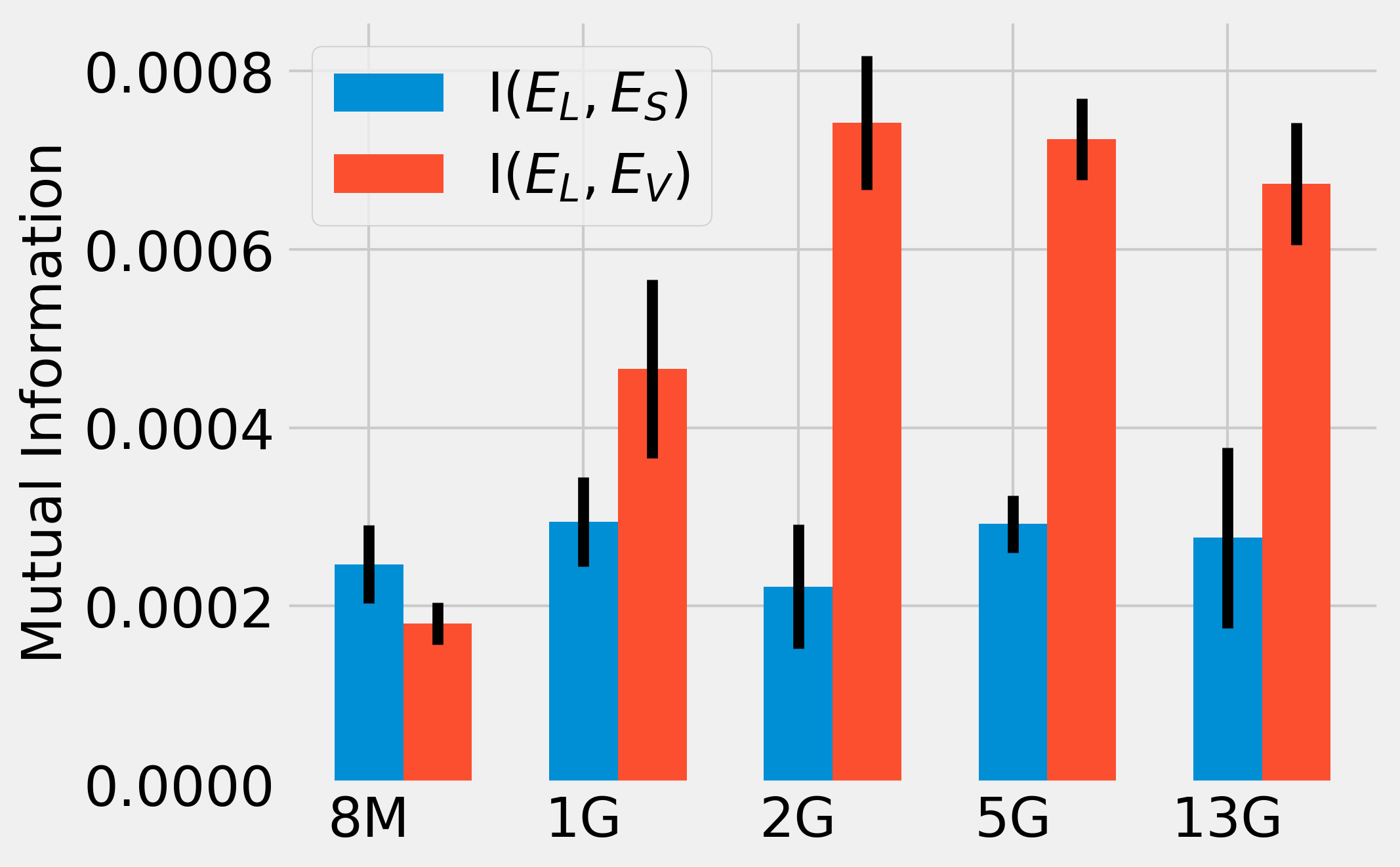}
\caption{$I_{\!H\!S\!I\!C}$, $\sigma$: median, $d = 50$}
\label{f:MI_HSIC_sigma-median_d50_quanities}
\end{subfigure}
\begin{subfigure}[t]{.49\linewidth}
\includegraphics[width=\textwidth]{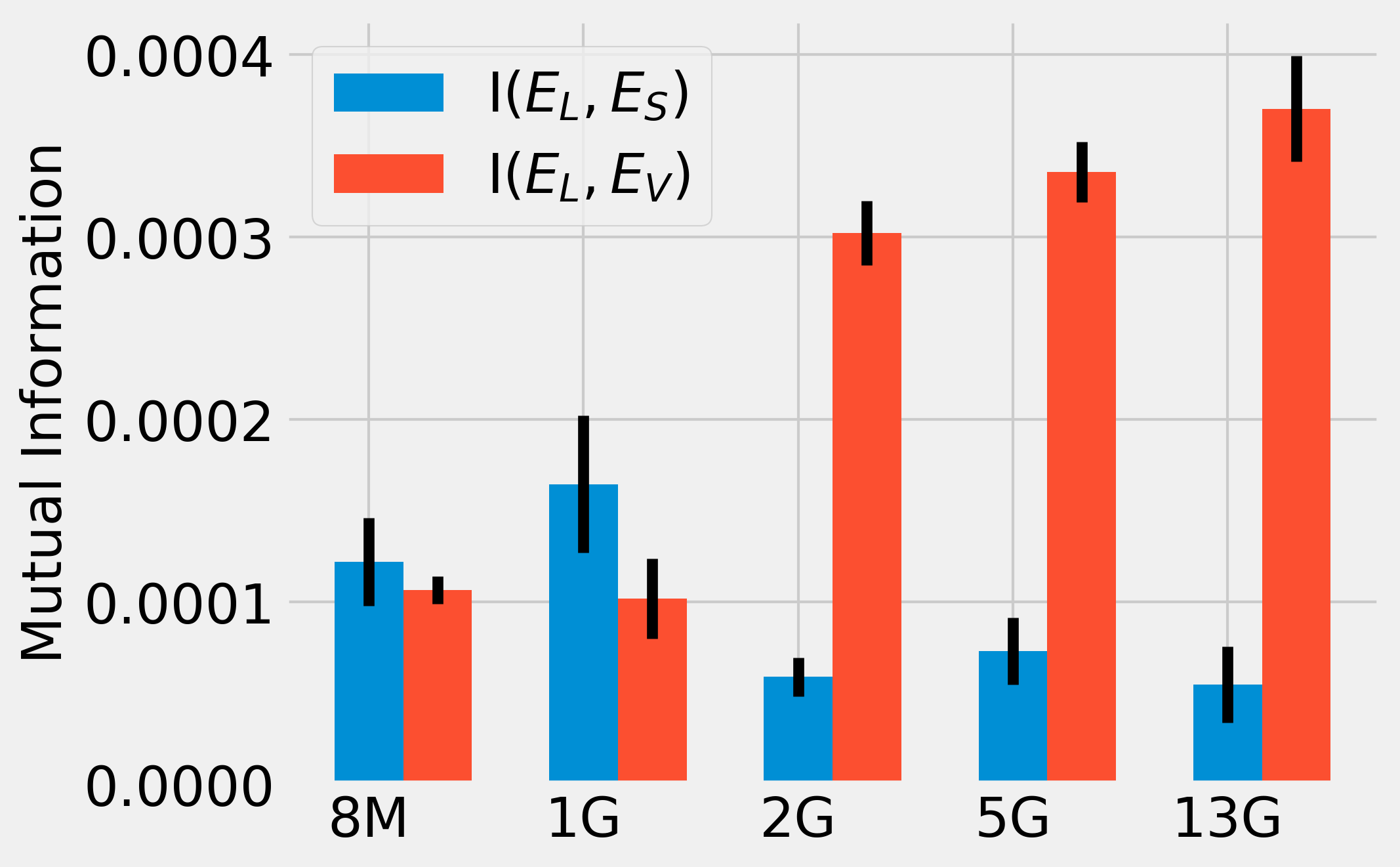}
\caption{$I_{\!H\!S\!I\!C}$, $\sigma$: median, $d\!=\!100$}
\label{f:MI_HSIC_sigma-median_d100_quanities}
\end{subfigure}
\caption{Estimated Mutual Informations: $I(E_L, E_V)$ (red) and $I(E_L, E_S)$ (blue) for different $E_L$ corpus sizes.}
\label{f:MI_quanities}
\end{figure}

\begin{figure}[t!]
\centering
\begin{subfigure}[t]{.49\linewidth}
\includegraphics[width=\textwidth]{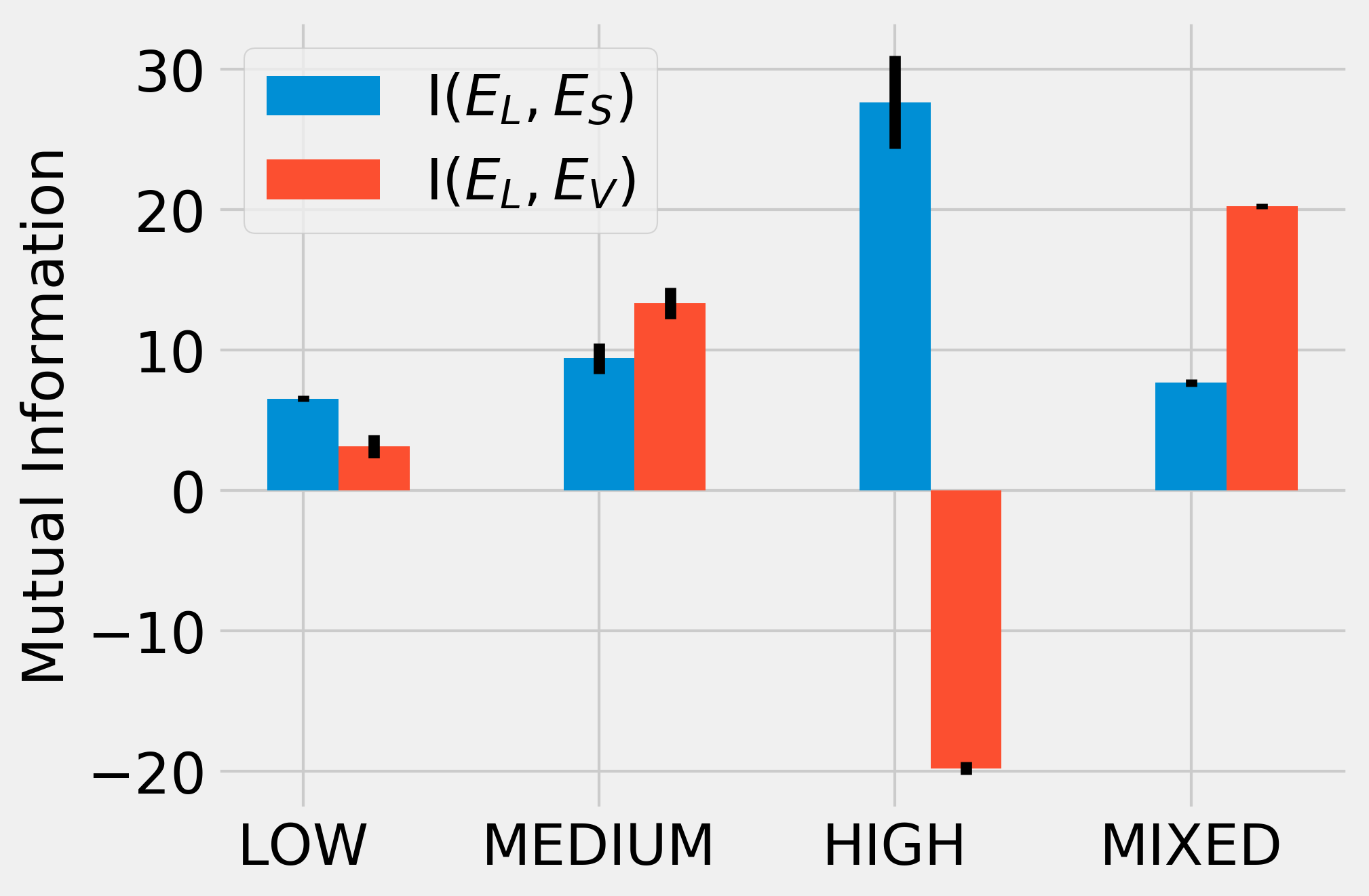}
\caption{$I_{\!K\!N\!N}$}
\label{f:MI_knn_freqranges}
\end{subfigure}
\begin{subfigure}[t]{.49\linewidth}
\includegraphics[width=\textwidth]{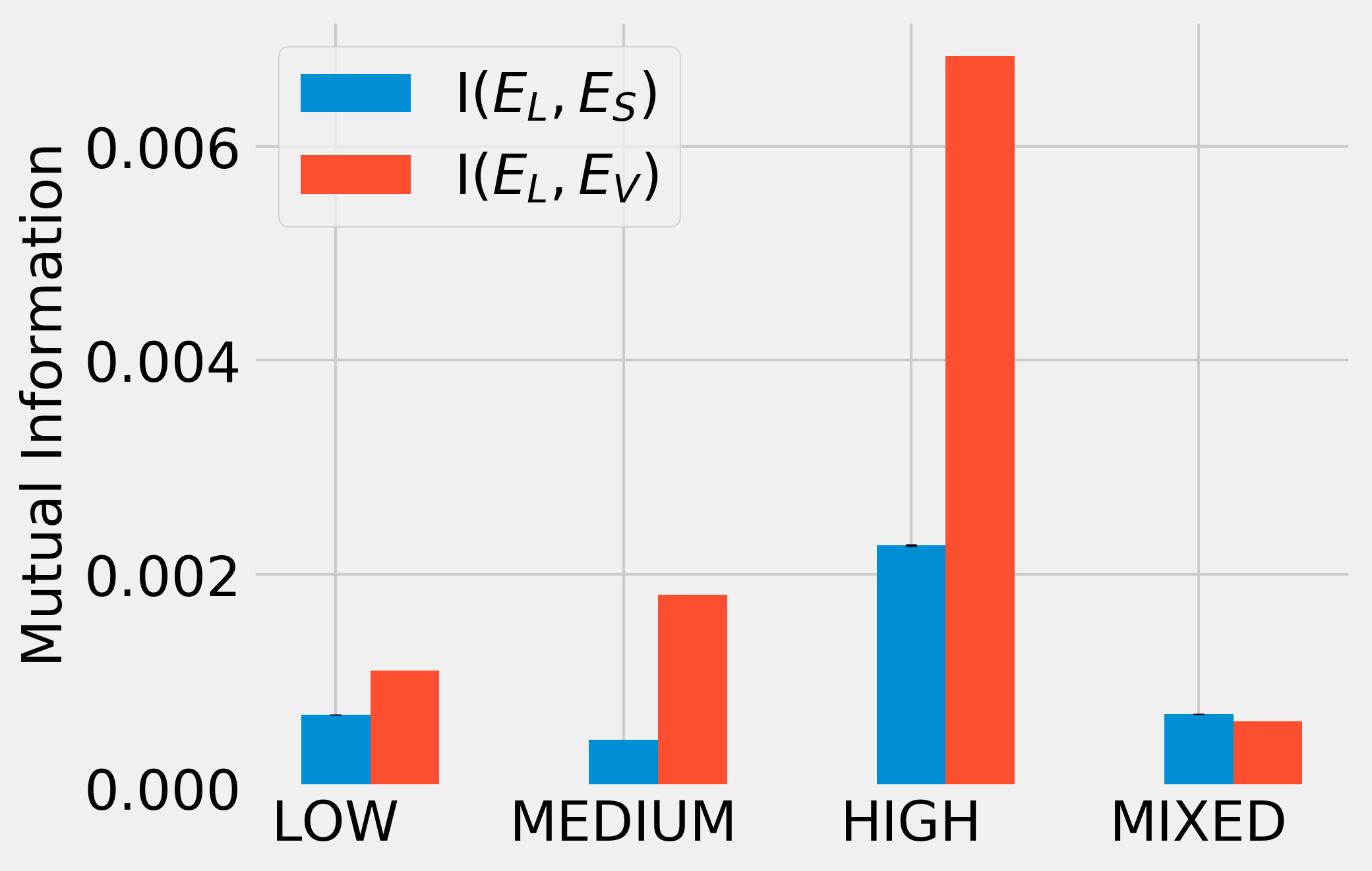}
\caption{$I_{\!H\!S\!I\!C}$, $\sigma = 1$, $d = max$}
\label{f:MI_HSIC_sigma1_freqranges}
\end{subfigure}
\begin{subfigure}[t]{.49\linewidth}
\includegraphics[width=\textwidth]{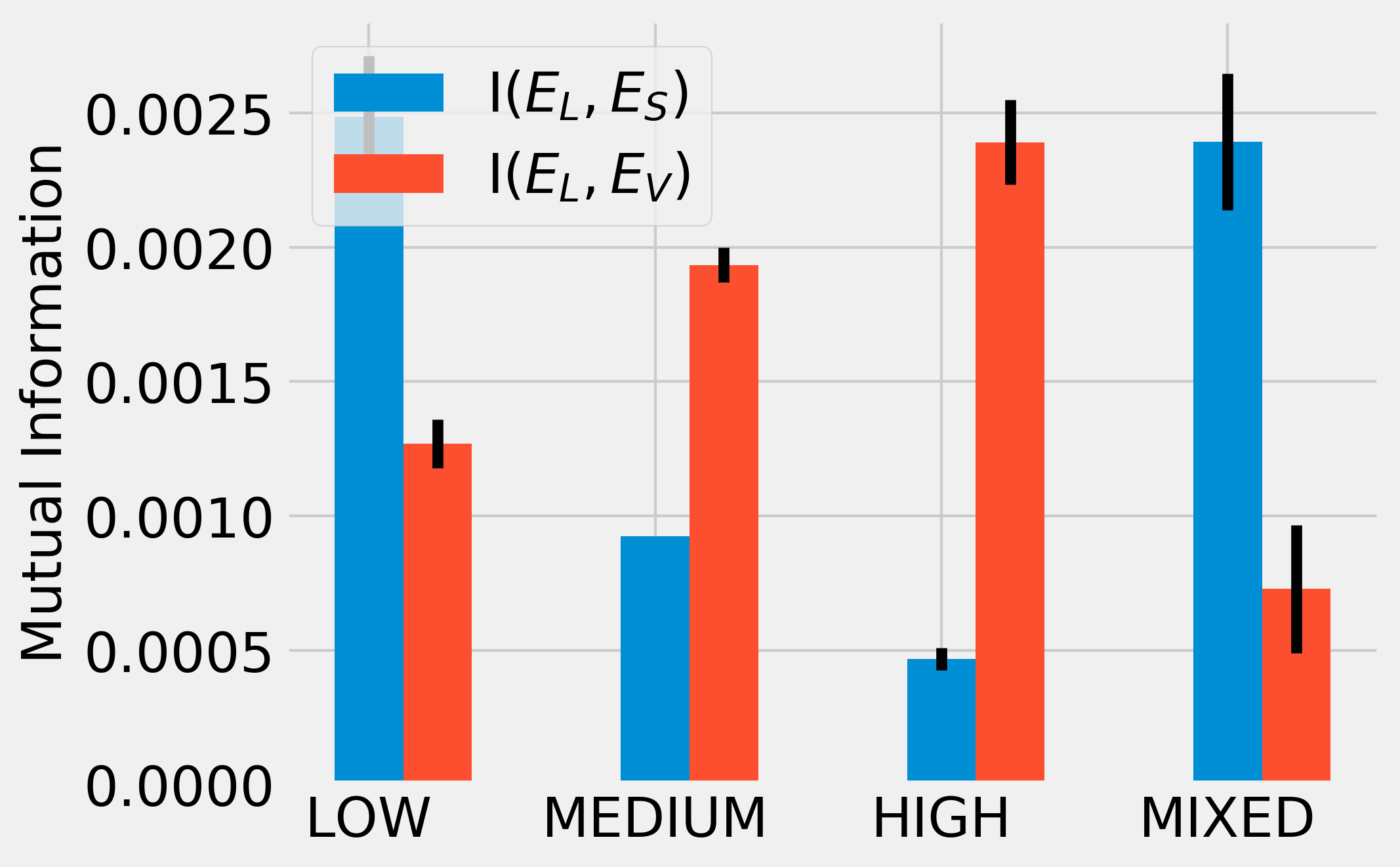}
\caption{$I_{\!H\!S\!I\!C}$, $\sigma$: median, $d = 3$}
\label{f:MI_HSIC_sigma-median_d3_quanities}
\end{subfigure}
\begin{subfigure}[t]{.49\linewidth}
\includegraphics[width=\textwidth]{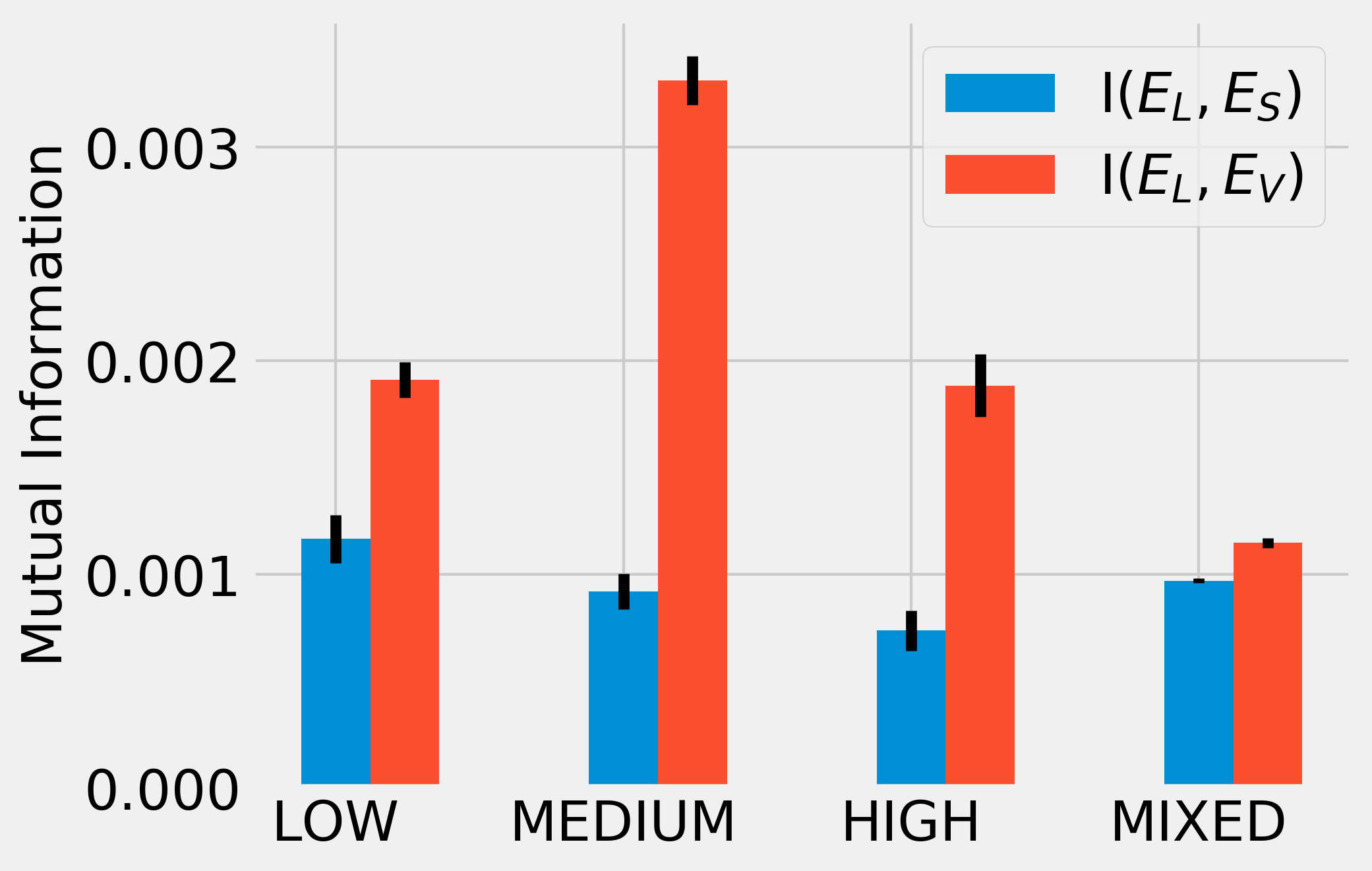}
\caption{$I_{\!H\!S\!I\!C}$, $\sigma$: median, $d = 10$}
\label{f:MI_HSIC_sigma-median_d10_freqranges}
\end{subfigure}
\begin{subfigure}[t]{.49\linewidth}
\includegraphics[width=\textwidth]{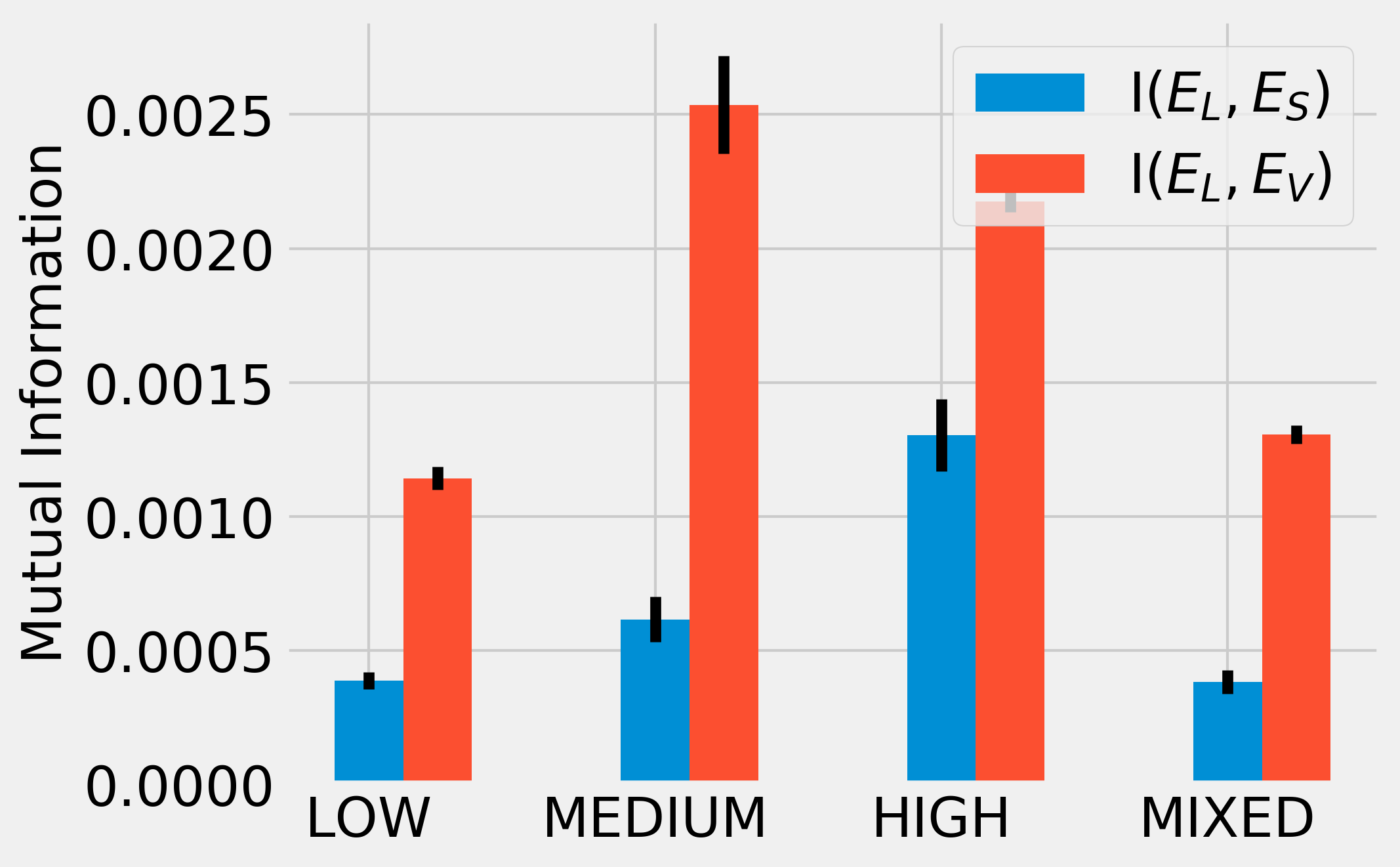}
\caption{$I_{\!H\!S\!I\!C}$, $\sigma$: median, $d = 13$}
\label{f:MI_HSIC_sigma-median_d13_freqranges}
\end{subfigure}
\begin{subfigure}[t]{.49\linewidth}
\includegraphics[width=\textwidth]{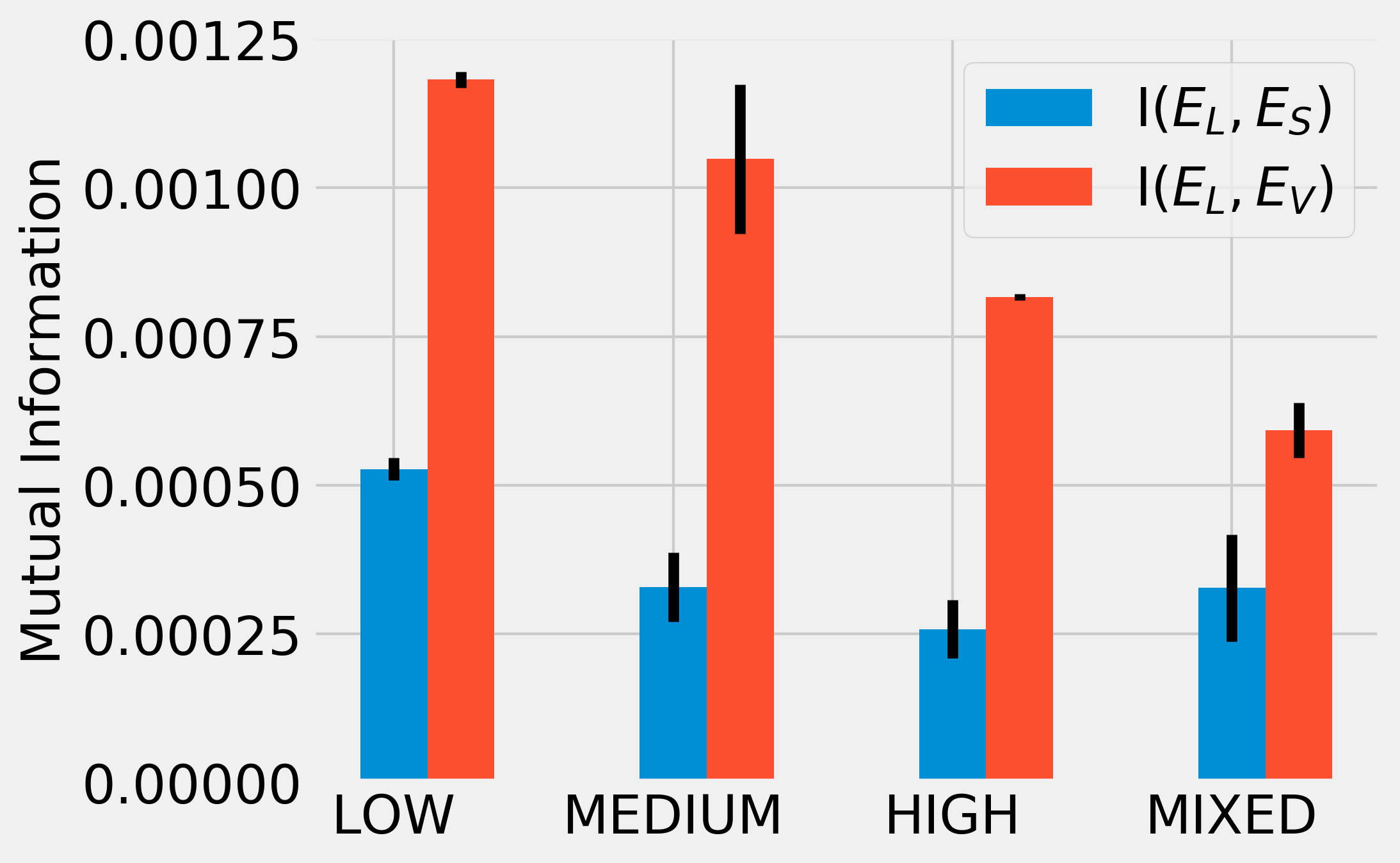}
\caption{$I_{\!H\!S\!I\!C}$, $\sigma$: median, $d = 50$}
\label{f:MI_HSIC_sigma-median_d50_freqranges}
\end{subfigure}
\begin{subfigure}[t]{.49\linewidth}
\includegraphics[width=\textwidth]{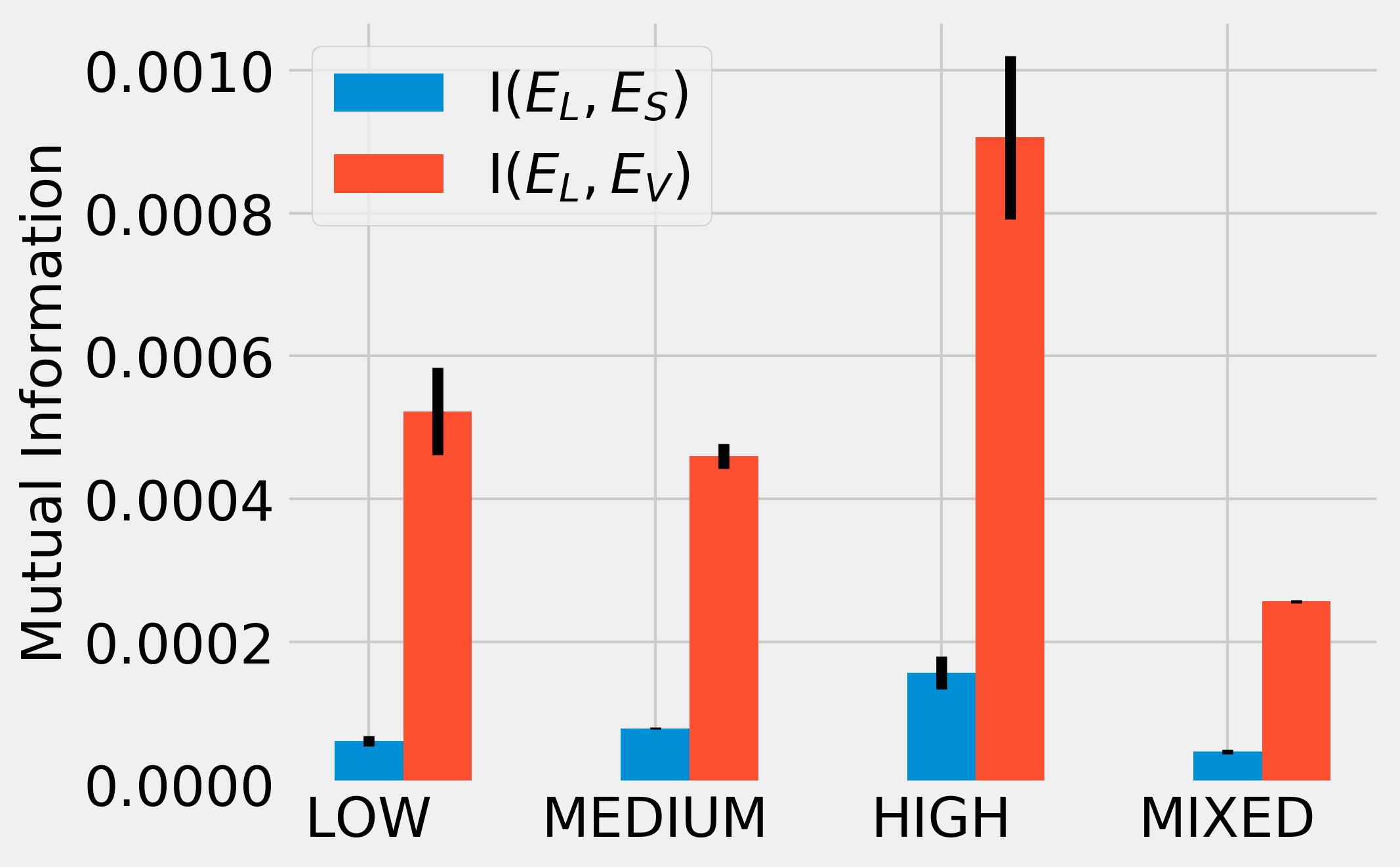}
\caption{$I_{\!H\!S\!I\!C}$, $\sigma$: median, $d\!=\!100$}
\label{f:MI_HSIC_sigma-median_d100_freqranges}
\end{subfigure}
\caption{Estimated Mutual Informations: $I(E_L, E_V)$ (red) and $I(E_L, E_S)$ (blue) for different word frequency ranges.}
\label{f:MI_freqranges}
\end{figure}

Results for different projected dimensions $d$ and $\sigma$ settings are shown in Fig.~\ref{f:MI_quanities} and \ref{f:MI_freqranges}.

\section{Cluster Structure} \label{app:cluster}

\begin{table*}
 \centering
 \setlength\tabcolsep{3pt}
 \begin{tabular}{l|l|l}
 \hline\hline
  \textbf{WordNet lb.} &   \textbf{Own lb.} & \textbf{Members}   \\
 \hline 
  \makecell[tr]{artifact\\line\\whole}                           &                   train & 
  \makecell[tl]{railway, railroad, subway, curve, tunnel, run, shelter, train, station, tram,\\
   highway, track, rail, way, engine, stop, gate, bridge, smoke}  \\
 \hline
  \makecell[tr]{plant organ\\plant\\vascular plant} & plants & 
  \makecell[tl]{bloom, foliage, grave, dead, vine, blossom, ivy, pod, cactus, tree, moss,\\ 
  root, leave, limb, forest, bush, plant, lily, branch, weed, leaf, vein, sunshine,\\ 
  log, fence, flower, sunlight, wood, palm, bench, sun}   \\
 \hline
  \makecell[tr]{structure\\area\\room} & room & 
  \makecell[tl]{classroom, hallway, hall, closet, bedroom, room, bathroom, garage, office, cafe,\\
 museum, doorway, kitchen, shop, restaurant, store, mannequin, stadium, market,\\
  ceiling, corner}  \\
 \hline
  \makecell[tr]{bird\\vertebrate\\person} &  \makecell[tl]{birds,\\animals} & 
  \makecell[tl]{hummingbird, gull, peacock, hawk, pelican, crow, parrot, seagull, wing, swan,\\
   pigeon, owl, goose, flamingo, nest, eagle, tail, bird, silhouette, duck, chest, body,\\
    ledge, giraffe, zebra}  \\
 \hline
  \makecell[tr]{food\\produce\\solid}                            &                    food & 
  \makecell[tl]{drizzle, nuts, herb, beef, flour, season, cereal, cherry, breakfast, sugar, steak, bacon, \\
  burger, butter, rice, meat, meal, sauce, dinner, pie, raspberry, lunch, sushi, bean,\\
  mustard, pepper, seed, salt, soup, cheese, tomato, hot, berry, potato, dessert,\\
  strawberry, salad, cardboard, food, bone, lemon, burn, frost, chocolate, bread,\\
  turkey, sandwich, spoon, pizza, chicken, shell, candy, peel, cooking, bubble,\\
  knife, fruit, fish, donut, cake, apple, ice, banana, orange}   \\
 \hline
 \end{tabular}
 \caption{Examples from the 20 K-means clusters in $E_S$.}
 \label{t:E_S_20_clusters}
\end{table*}

\begin{table*}
 \centering
 \setlength\tabcolsep{3pt}
 \begin{tabular}{l|l|l}
 \hline\hline
 \textbf{WordNet lb.} &   \textbf{Own lb.} & \textbf{Members}\\
\hline
 \makecell[tr]{food\\nutriment\\foodstuff}                          &                    food & 
 \makecell[tl]{butter, cheese, bread, chicken, soup, sauce, dessert, beef, salad, meat, cake,\\
  steak, tomato, potato, pizza, flour, milk, meal, vinegar, bacon, pie, cooking\\
  sushi, sandwich, breakfast, burger, menu}  \\
\hline
 \makecell[tr]{vascular plant\\plant organ\\plant part}             &                   plants & 
 \makecell[tl]{flower, flowers, tree, blossom, dandelion, foliage, fruit, weed, cactus, lily,\\
 bloom, shade, leaf, grass, sunflower, poppy, vine, plant, garden, iris, grow,\\
 daisy, oak, bulb, rust, herb, moss, tulip, palm, maple, root, tall, bush, seed, family}  \\
\hline
 \makecell[tr]{artifact\\structure\\whole}                          & \makecell[tl]{classical \\architecture} & 
 \makecell[tl]{tower, building, marble, staircase, fountain, doorway, roof, chapel, steeple\\
 porch, ceiling, mural, glass, wall, brick, statue, stone, arch, monument, dome,\\
 window, gravestone, sculpture, aisle, tiles, gate, interior, painted, decoration,\\
 concrete, church, graveyard, cathedral, curtain, painting, palace, clock, grave,\\
 portrait, choir, architecture, pyramid, memorial, square, castle, skyscraper,\\
 museum, cemetery, temple, organ}  \\
\hline
 \makecell[tr]{body part\\part\\artifact}                           &                    body parts & 
 \makecell[tl]{skin, spine, neck, bone, chest, throat, shoulder, wrist, stomach, ear, jaw, cheek,\\
 lips, nose, eyes, eye, limb, toe, belly, skull, abdomen, finger, teeth, elbow, cord,\\
 whiskers, knee, thumb, tooth, muscle, ankle, tail, paws, lip, brain, flesh, leg, body,\\
 calf, heart, blood, tongue, brow, pain, tear, blade, mouth, liver, gut, arm, marrow,\\
 curled, canine, feathers, foot, vein, hip, cancer} \\
\hline
 \makecell[tr]{change\\act\\be}                                     &                    verbs & 
 \makecell[tl]{bring, get, come, want, go, keep, take, know, find, say, give, make, understand,\\
 put, listen, enjoy, feel, leave, think, learn, imagine, gather, believe, fail, arrange,\\
 add, lose, create, way, hear, send, meet, collect, carry, avoid, buy, remain, allow,\\
 appear, might, enter, arrive, seem, entertain, break, steal, receive, stop, stand,\\
 build, locked, compare, retain, sell, handle, danger, eat, wander, face, unhappy,\\
 protect, please, pray, become, walk, expand, travel, plenty, greet, inspect, comfort,\\
 huge, possess, dominate, attach, roam, participate, speak, step, drawn, construct,\\
 replace, divide, great, living}  \\
\hline
\end{tabular}
\caption{Examples from the 20 K-means clusters in $E_L$.}
\label{t:E_L_20_clusters}
\end{table*}

\begin{table*}
 \centering
 \setlength\tabcolsep{3pt}
 \begin{tabular}{l|l|l}
 \hline\hline
 \textbf{WordNet lb.} &   \textbf{Own lb.} & \textbf{Members}\\
 \hline 
 \makecell[tr]{bird\\aquatic bird\\seabird}                              &                   birds & seagull, gull, goose, duck, pelican, swan, mallard, stork, eagle, flamingo  \\
\hline 
 \makecell[tr]{vascular plant\\plant\\grow}                              &                    plants & 
 \makecell[tl]{weed, bunch, maple, cancer, iris, poppy, dandelion, leave, flower, rose, foliage,\\
 grow, plant, cactus, spring, tulip, ivy, palm, lily, leaf, daisy, tree, root, wheat, wool,\\
 raspberry, tobacco, flowers, blossom, butterfly, sunflower, cotton, herb, violet, oak,\\
 moss, strawberry, nest, dew, berry, rice, branch, coal}  \\
\hline
 \makecell[tr]{food\\nutriment\\substance}                               &                    food & 
 \makecell[tl]{sushi, meal, sandwich, pie, breakfast, lunch, food, supper, flour, cereal, sweet,\\
 dessert,mdinner, subway, diet, cake, date, steak, sauce, bread, copper, nuts, bacon,\\
 cooking, beef, meat, bakery, knitting, eat, potato, salad, donut, pizza, burger, coffee,\\
 soup, bean, cheese, vitamin, fruit, pumpkin, rock, marrow, market, timber}  \\
\hline
 \makecell[tr]{artifact\\change\\cover}                                  &                    
 \makecell[tl]{colours,\\materials} & 
 \makecell[tl]{texture, fabric, cloth, metal, rain, concrete, paper, suds, rough, words, stone, wall,\\
  square, dense, leather, quote, wood, frost, mud, noise, text, purple, carpet, blue, tiles,\\
  dirt, droplets, red, sand, fog, formula, mist, pattern, handwriting, green, straw, linen,\\
  asphalt, stripes, crowd, marble, yellow, black, brown, grey, grass, white}  \\
\hline
 \makecell[tr]{travel\\change\\object}                                   &                    vacation & 
 \makecell[tl]{island, view, reflection, harbor, nice, side, sea, summer, tropical, pollution, port,\\
  aircraft, pier, travel, surfers, journey, sunny, coast, flying, morning, ocean, seashore,\\
  horizon, mare, holiday, lake, surf, shore, vacation, bay, airport, cliff, sunlight, air,\\
  river, storm, ship, fishing, beach, desert, harbour, puddle, flight, sailing, evening,\\
  sunrise, skyline, vessel, lighthouse, dawn, sunset, rocket, mountain, whale,\\
  underwater, boat, swimming, swim, plane, dusk, jet, cloud, sky, airplane, ski} \\
\hline
\end{tabular}
\caption{Examples from the 20 K-means clusters in $E_V$.}
\label{t:E_V_20_clusters}
\end{table*}

Example clusters on the embeddings' common vocabulary can be found in Tbl.~\ref{t:E_S_20_clusters}-\ref{t:E_V_20_clusters}. Tbl.~\ref{t:jaccard_sim_nums} includes an analysis on Jaccard similarities among different clusterings using K-means and Agglomerative clustering. All clusters with automatic WordNet, as well as the authors' annotation will be publicly available online.

Main observations:
\begin{itemize}
\item Wikipedia based $E_L$ has more clusters with abstract topics, such as verbs, activities and communication. 
\item $E_S$ has more concrete clusters e.g., train, vehicles, building structures, containers or furnishing. 
\item Image based $E_V$ includes more clusters related to the outdoors, such as ``travel'', ``transportation'', ``landscape'' and ``vacation'' and on appearance, such as ``colours \& materials''.
\item Concepts that all three embeddings capture: ``food'', ``colours'', ``plants'', ``animals'' and ``body parts''.
\end{itemize}

\begin{table}
    \centering
    \setlength\tabcolsep{1.5pt}
    \begin{tabular}{l| cc | cc | cc | ll}
    \multicolumn{9}{c}{\textbf{Cross-modalities}} \\
    \hline\hline
    & \multicolumn{2}{c|}{\textbf{$>$0.2}} & \multicolumn{2}{c|}{\textbf{$>$0.3}} & \multicolumn{2}{c|}{\textbf{$>$0.4}} & \multicolumn{2}{c}{\textbf{Max}} \\
    Clust. & K-m & Agg & K-m & Agg & K-m & Agg & K-m & Agg\\
    \hline
    $E_S$-$E_L$ & 12 & 6 & 1 & 1 & 0 & 0 & 0.358 & 0.347\\
    $E_S$-$E_V$ & 8  & 9 & 2 & 4 & 0 & 2 & 0.363 & 0.5\\
    $E_L$-$E_V$ & 9  & 9 & 5 & 5 & 4 & 2 & 0.479 & 0.467\\
    \hline
    \end{tabular}

    \setlength\tabcolsep{2.5pt}
    \begin{tabular}{l| c c c c c c | l}
    \multicolumn{8}{c}{\textbf{K-means -- Agglomerative}} \\
    \hline\hline
    & \textbf{$>$0.2} & \textbf{$>$0.3} & \textbf{$>$0.4} & \textbf{$>$0.5} & \textbf{$>$0.6} & \textbf{$>$0.7} & \textbf{Max}\\\hline
    $E_L$-$E_L$ & 18 & 14 & 9 & 4 & 2 & 1 & 0.79\\
    $E_S$-$E_S$ & 16 & 15 & 13 & 9 & 5 & 1 & 0.729\\
    $E_V$-$E_V$ & 23 & 16 & 13 & 8 & 4 & 0 & 0.644\\
    \hline
    \end{tabular}

    \caption{Number of cluster pairs out of $20^2$ with Jaccard similarities above thresholds of [0.2,\dots, 0.7]. Last column shows the maximum similarity.}
    \label{t:jaccard_sim_nums}
\end{table}

\end{document}